%% file: main.tex
\pdfoutput=1

\documentclass[10pt,twocolumn,letterpaper]{article}

\usepackage{iccv}
\usepackage{times}
\usepackage{epsfig}
\usepackage{graphicx}
\usepackage{amsmath}
\usepackage{amssymb}
\usepackage{enumitem}

\usepackage{multirow}
\usepackage{pifont}
\usepackage{makecell}
\usepackage{wrapfig}

\usepackage[accsupp]{axessibility}  

\usepackage[pagebackref=true,breaklinks=true,letterpaper=true,colorlinks,bookmarks=false]{hyperref}

\usepackage{xcolor}
\usepackage[capitalize]{cleveref}
\crefname{section}{Sec.}{Secs.}
\Crefname{section}{Section}{Sections}
\Crefname{table}{Table}{Tables}
\crefname{table}{Tab.}{Tabs.}

\newcommand{\fulltaskname}{Panoptic Scene Graph Generation from Purely Textual Descriptions\xspace}
\newcommand{\taskname}{Caption-to-PSG\xspace}
\newcommand{\modelname}{TextPSG\xspace}

\newcommand{\moduleAA}{{{region grouper}}}
\newcommand{\moduleBB}{{{entity grounder}}}
\newcommand{\moduleCC}{{{segment merger}}}
\newcommand{\moduleDD}{{{label generator}}}

\newlength\savewidth
\newcommand\shline{\noalign{\global\savewidth\arrayrulewidth
  \global\arrayrulewidth 1pt}\hline\noalign{\global\arrayrulewidth\savewidth}}

\makeatletter
\newcommand*\bigcdot{\mathpalette\bigcdot@{.7}}
\newcommand*\bigcdot@[2]{\mathbin{\vcenter{\hbox{\scalebox{#2}{$\m@th#1\bullet$}}}}}
\makeatother

\iccvfinalcopy 

\def\iccvPaperID{3290} 

\ificcvfinal\fi

\begin{document}

\title{TextPSG: Panoptic Scene Graph Generation from Textual Descriptions}

\author{
Chengyang Zhao\textsuperscript{ 1} \quad 
Yikang Shen\textsuperscript{ 2} \quad
Zhenfang Chen\textsuperscript{ 2} \\
Mingyu Ding\textsuperscript{ 3} \quad 
Chuang Gan\textsuperscript{ 2,4} \\
\textsuperscript{1}Peking University \quad
\textsuperscript{2}MIT-IBM Watson AI Lab \\
\textsuperscript{3}UC Berkley \quad
\textsuperscript{4}UMass Amherst
}

\maketitle
\ificcvfinal\thispagestyle{empty}\fi

\input{tex/00_abstract.tex}

\input{tex/01_introduction.tex}

\input{tex/02_related_work.tex}

\input{tex/03_problem_formulation.tex}

\input{tex/04_method.tex}

\input{tex/05_experiments.tex}

\input{tex/06_conclusion.tex}

{\small
\bibliographystyle{ieee_fullname}
\bibliography{egbib}
}

\clearpage
\appendix
\input{tex/ap_01_method.tex}

\input{tex/ap_02_experiment.tex}

\input{tex/ap_03_result_analysis.tex}

\end{document}

%% file: tex/00_abstract.tex
\begin{abstract}
Panoptic Scene Graph has recently been proposed for  comprehensive scene understanding.
However, previous works adopt a fully-supervised learning manner, requiring large amounts of pixel-wise densely-annotated data, which is always tedious and expensive to obtain. 
To address this limitation, we study a new problem of~\fulltaskname (\taskname).
The key idea is to leverage the large collection of free image-caption data on the Web alone to generate panoptic scene graphs. 
The problem is very challenging for three constraints: 1) no location priors; 2) no explicit links between visual regions and textual entities; and 3) no pre-defined concept sets. 
To tackle this problem, we propose a new framework \modelname consisting of four modules, \ie, a \moduleAA, an \moduleBB, a \moduleCC, and a \moduleDD, with several novel techniques.
The \moduleAA~first groups image pixels into different segments and the \moduleBB~then aligns visual segments with language entities based on the textual description of the segment being referred to.
The grounding results can thus serve as pseudo labels enabling the \moduleCC~to learn the segment similarity as well as guiding the \moduleDD~to learn object semantics and relation predicates, resulting in a fine-grained structured scene understanding. 
Our framework is effective, significantly outperforming the baselines and achieving strong out-of-distribution robustness. We perform comprehensive ablation studies to corroborate the effectiveness of our design choices and provide an in-depth analysis to highlight future directions.
Our code, data, and results are available on our project page: \href{https://textpsg.github.io/}{https://textpsg.github.io/}.

\end{abstract}

%% file: tex/01_introduction.tex
\section{Introduction}

A scene graph is a directed-graph-based abstract representation of the objects and their relations within a scene. 
It has been widely utilized to develop a structured scene understanding of object semantics, locations, and relations, which
facilitates a variety of downstream applications, such as image generation~\cite{johnson_2018_image, dhamo_2020_simsg}, visual reasoning~\cite{teney_2017_graph, aditya_2017_graph, shi_2019_explainable}, and robotics~\cite{Amiri_2022_ReasoningWS, Gadre_2022_ContinuousSR}.

\input{figure/01_teaser.tex}

Typically, in a scene graph, each node denotes an object in the scene located by a bounding box (bbox) with a semantic label, and each directed edge denotes the relation between a pair of objects with a predicate label.
Nonetheless, a recent work~\cite{yang_2022_psg} points out that such a bbox-based form of scene graph is not ideal enough. Firstly, compared with pixel-wise segmentation masks, bboxes are less fine-grained and may contain some noisy pixels belonging to other objects, limiting the applications for some downstream tasks.
For example, as shown in~\cref{fig:teaser} (a), about half of the pixels in the yellow bbox for \textit{girl} belong to \textit{wall}.
Secondly, it is challenging for bboxes to cover the entire scene without ambiguities caused by overlaps, which prevents a scene graph from including every object in the scene for a complete description.
To this end, the work~\cite{yang_2022_psg} proposes the concept of \textit{Panoptic Scene Graph} (PSG), in which each object is grounded by a panoptic segmentation mask, to reach a comprehensive structured scene representation.

However, all existing works~\cite{yang_2022_psg, Wang_2023_PSGWorkshop} approach PSG generation through a fully-supervised manner, \ie, learning to perform panoptic segmentation and relation prediction from manually-annotated datasets with explicit supervision for both segmentation and relation prediction. Unfortunately, it is extremely labor-intensive to build such datasets, making it difficult to scale up to cover more complex scenes, object semantics, and relation predicates, thus significantly limiting the generalizability and the application of these methods to the real world. For instance, the current PSG dataset~\cite{yang_2022_psg} only covers 133 object semantics and 56 relation predicates.

To relieve the reliance on densely-annotated data, weakly-supervised methods~\cite{Ye_2021_Linguistic, zhong_2021_SGGNLS, Li_2022_IntegratingOA} for scene graph generation are promising. 
These methods could induce scene graphs from image-caption pairs, which can be easily harvested from the Web for free. 
Even so, they still rely heavily on two strong preconditions, \ie, a powerful region proposal network (\eg,~\cite{ren_2015_fasterrcnn}) and a pre-defined set of object semantics and relation predicates.
Although these preconditions facilitate the learning process of the methods, they also limit the generalizability for locating novel objects (unforeseen objects for the region proposal network) and constrain the understanding into the limited concept set.

Inspired by previous weakly-supervised methods, we introduce a new problem, \textit{\fulltaskname} (\taskname), to explore a holistic structured scene understanding without labor-intensive data annotation. Considering the limitation of the preconditions mentioned, we set three constraints to \taskname~to reach a more comprehensive and generalizable understanding, which results in a very challenging problem: a) only image-caption pairs are provided during training, without any location priors in either region proposals or location supervision; b) the explicit links between regions in images and entities in captions are missing; c) no concept sets are pre-defined, \ie, neither object semantics nor relation predicates are known beforehand.

Given these three constraints, we argue that there are two key challenges for the model to solve the problem. 
Firstly, the model should learn to ground entities in language onto the visual scene without explicit location supervision, \ie, the ability to perform partitioning and grounding, as shown in~\cref{fig:teaser} (b), should be developed purely from textual descriptions.
Secondly, during training, the model should also learn the object semantics and relation predicates from textual descriptions, as shown in~\cref{fig:teaser} (c), without pre-defined fixed object and relation vocabularies.
By solving these challenges, the model could associate visual scene patterns with textual descriptions,
gradually acquire common sense among them, and finally reach a more comprehensive and generalizable understanding, including novel object location, extensive semantics recognition, and complex relation analysis, which is more suitable to the real world.

With these considerations, we propose a novel framework, \modelname, as the first step towards this challenging problem. \modelname~consists of a series of modules to cooperate with each other, \ie, a \moduleAA, an \moduleBB, a \moduleCC, and a \moduleDD.
The \moduleAA~learns to merge image regions into several segments in a hierarchical way based on object semantics, similar to~\cite{xu_2022_group}.
The \moduleBB~employs a fine-grained contrastive learning strategy~\cite{yao_2022_filip} to bridge the textual description and the visual content, grounding entities in the caption onto the image segments.
With the entity-grounding results as pseudo labels, the \moduleCC~learns similarity matrices to merge small image segments during inference, while the \moduleDD~learns the prediction of object semantics and relation predicates. 
Specifically, in the \moduleCC, we propose to leverage the grounding as explicit supervision for merging, compared with~\cite{xu_2022_group} which learns merging in a fully implicit manner, to improve the ability of location.
In the \moduleDD, different from all previous pipelines for scene graph generation, we reformulate the label prediction as an auto-regressive generation problem rather than a classification problem, and employ a pre-trained language model~\cite{li_2022_blip} as the decoder to leverage the pre-learned common sense. We further design a novel prompt-embedding-based technique (PET) to better incorporate common sense from the language model. 
Our experiments show that~\modelname~significantly outperforms the baselines and achieves strong out-of-distribution (OOD) robustness. Comprehensive ablation studies corroborate the effectiveness of our design choices. 
As a side product, the proposed grounder and merger modules also have been observed to enhance text-supervised semantic segmentation.

In spite of the promising performance of~\modelname, certain challenges persist. We delve into an in-depth analysis of the failure cases, provide a model diagnosis, and discuss potential future directions for enhancing our framework.

To sum up, our contributions are as follows:

\begin{itemize}[align=right,itemindent=0em,labelsep=2pt,labelwidth=1em,leftmargin=*,itemsep=0em] 
\item We introduce a new problem,~\textit{\fulltaskname} (\taskname), to alleviate the burden of human annotation for PSG by learning purely from the weak supervision of captions.

\item We propose a new modularized framework,~\modelname, with several novel techniques, which significantly outperforms the baselines and achieves strong OOD robustness. 
We demonstrate that the proposed modules in~\modelname can also facilitate text-supervised semantic segmentation.

\item We perform an in-depth failure case analysis with a model diagnosis, and further highlight future directions.
\end{itemize}

%% file: figure/01_teaser.tex
\begin{figure}[t]
  \centering
  \includegraphics[width=1\linewidth]{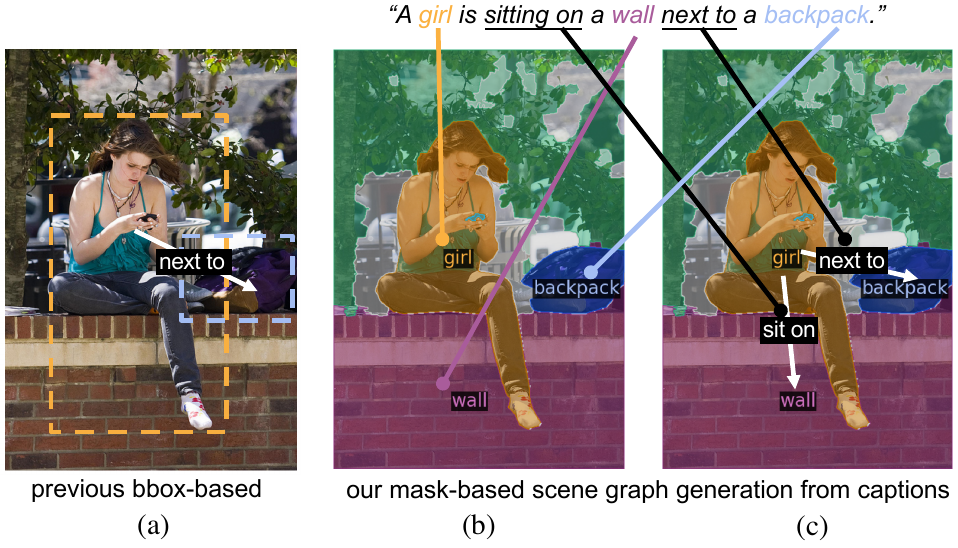}
   \caption{
   \textbf{Problem Overview.} 
   Different from the traditional bbox-based form of the scene graph as shown in (a), \taskname aims to generate the mask-based panoptic scene graph. In \taskname, the model has no access to any location priors, explicit region-entity links, or pre-defined concept sets. Consequently, the model is required to learn partitioning and grounding as illustrated in (b), as well as object semantics and relation predicates as illustrated in (c), all purely from textual descriptions.
   }
   \label{fig:teaser}
\end{figure}

%% file: tex/02_related_work.tex
\section{Related Work\label{sec:related_work}}

\noindent\textbf{Bbox-based Scene Graph Generation.}
Bbox-based scene graph generation aims to create a structured representation of object semantics, locations, and relations in the scene, where each object is identified by a bbox. Most of existing works~\cite{Yang_2018_GraphRCNN, xu_2017_messagepass, Tang_2020_unbias, Gu_2019_external, Lin_2020_gpsnet} follow a fully-supervised approach to learn the generation from densely-annotated datasets~\cite{Krishna_2016_VG, hudson_2019_gqa}, which requires significant human labors. To reduce the labeling effort, some weakly-supervised methods have been proposed~\cite{Peyre_2017_WSVR, Zareian_2020_WSVSP, zhang_2017_ppr, Shi_2021_ASB}. Recent works~\cite{Ye_2021_Linguistic, zhong_2021_SGGNLS, Li_2022_IntegratingOA} further explore learning scene graph generation from image-caption pairs. However, they all rely on off-the-shelf region proposal networks for the location of objects in the scene, which are typically pre-trained on pre-defined fixed sets of object semantics, limiting their generalizability to locating unforeseen objects. To reach a more granular and accurate grounding,~\cite{Khandelwal_2021_SegmSG} proposes to ground each object by segmentation. A recent work~\cite{yang_2022_psg} further introduces the concept of PSG, where each object is identified by a panoptic segmentation mask, as a more comprehensive scene representation.

\noindent\textbf{Text-supervised Semantic Segmentation (TSSS).}
TSSS~\cite{xu_2022_group,li2022language,luddecke2022image, dong2022maskclip,luo2022segclip, zhou2022extract} aims to learn image pixel semantic labeling from image-caption pairs without fine-grained pixel-wise annotations.
Similar to TSSS, our proposed \taskname aims to learn to connect visual regions and textual entities from only image-caption pairs and has the potential to leverage the large collection of free data on the Web. 
However, different from TSSS, \taskname further requires the model to learn the relations among different visual regions, resulting in a higher-order structured understanding of visual scenes. In addition to unknown object semantics, \taskname does not assume any pre-defined relation predicate concepts.

\noindent\textbf{Visual Grounding.} 
Our work is also related to visual grounding~\cite{kazemzadeh2014referitgame,zeng2020dense,mao2016generation,gan2017vqs,zfchen2021iclr,plummer2015flickr30k}, which grounds entities in language onto objects in images. Early works~\cite{yu2016modeling,yu2018mattnet,chen2020cops} on visual grounding typically detect object proposals~\cite{ren_2015_fasterrcnn,uijlings_2013_selective} from images first and then match them with language descriptions by putting features of both modalities into the same feature space, which are in a fully-supervised learning manner. There are also some weakly-supervised grounding methods~\cite{karpathy2015deep,rohrbach2016grounding,chen19acl} which relieve the need for dense regional annotations by multiple instance learning~\cite{karpathy2015deep} or learning to reconstruct~\cite{rohrbach2016grounding}. Different from them, \taskname is more challenging since it requires grounding fine-grained object relations between entities without region proposal networks for a pre-defined object vocabulary.

%% file: tex/03_problem_formulation.tex
\input{figure/02_pipeline.tex}

\section{Problem Formulation\label{sec:problem}}
\noindent{\textbf{{\fulltaskname~(\taskname).}}
A PSG $\mathcal{G}=(\mathcal{V},\mathcal{E})$ is a directed graph representation of the objects and the relations among them in a scene image $I \in \mathbb{R}^{H \times W \times 3}$. Each node $v_i \in \mathcal{V}$ denotes an object in $I$ located by a panoptic segmentation mask $m_i \in \{0,1\}^{H \times W}$ with an object semantic label $o_i \in \mathcal{C}_o$, and each directed edge $e_{ij} \in \mathcal{E}$ denotes a pair of subject $o_i$ and object $o_j$ with a relation predicate label $r_{ij} \in \mathcal{C}_r$, where $\mathcal{C}_o$ and $\mathcal{C}_r$ are the defined concept sets of object semantics and relation predicates. Note that for a PSG, it is constrained that all segmentation masks could not overlap, \ie, $\sum_{i=1}^{|\mathcal{V}|} {m_i} \leq \mathbf{1}^{H \times W}$.

Given a large collection of paired scene images and textual descriptions $\mathcal{S} = \{(I_i, T_i)\}_i$, \taskname aims to learn PSG generation from purely text descriptions for a holistic structured scene understanding, \ie, during training, only $\mathcal{S}$ is provided as supervision, while during inference, with a scene observation $I'$ as input, the model is required to generate a corresponding PSG $\mathcal{G}'$.

\noindent{\textbf{{Three Constraints.}}
Note that in~\taskname, three important constraints are set to reach a more comprehensive and generalizable scene understanding: a) no location priors: different from all previous scene graph generation methods, neither pre-trained region proposal networks nor location supervision are allowed; b) no explicit region-entity links: the links between regions in the image $I$ and entities in the textual description $T$ are not provided; c) no pre-defined concept sets: the target concept sets defined for inference,~\ie, object semantics $\mathcal{C}_o$ and relation predicates $\mathcal{C}_r$, are unknown during training.

%% file: figure/02_pipeline.tex
\begin{figure*}[t]
  \centering
  \includegraphics[width=1\linewidth]{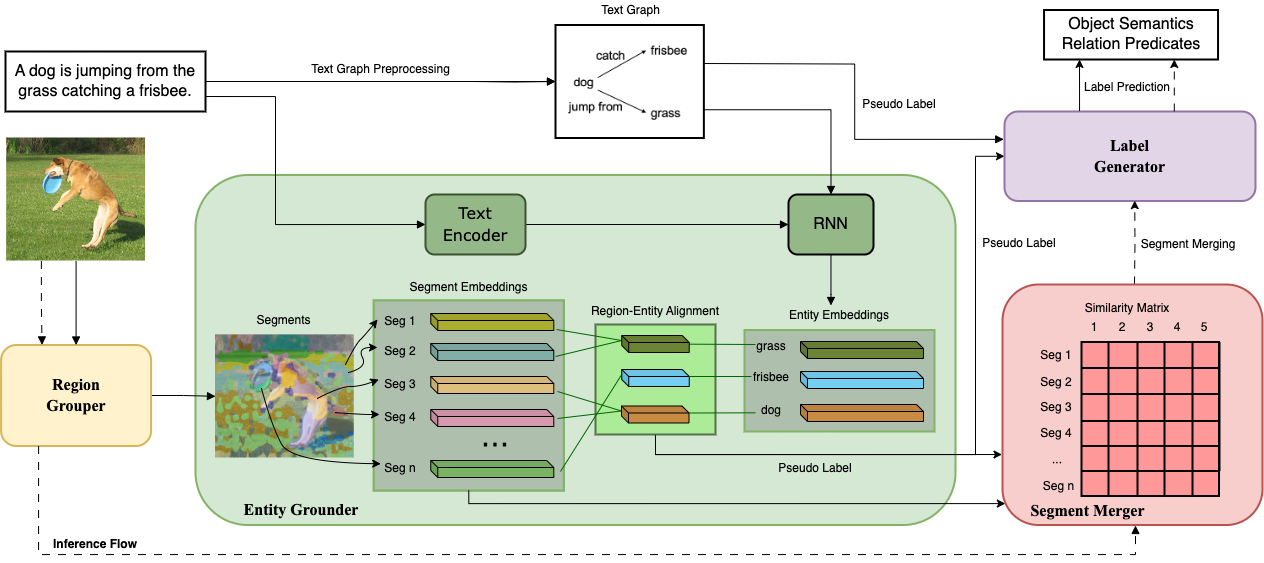}
   \caption{\textbf{Framework Overview of \modelname.} 
   The framework consists of four modules cooperating with each other: a~\moduleAA~to merge regions in the input image into several segments, an~\moduleBB~to ground entities in the caption onto the image segments, a~\moduleCC~to learn similarity matrices to merge small image segments during inference, and a~\moduleDD~to learn the prediction of object semantics and relation predicates. The solid arrows indicate the training flow, while the dash arrows indicate the inference flow. The arrows from the~\moduleAA~to the~\moduleDD~indicating the segment feature and mask query are omitted.
   }
   \label{fig:pipeline}
\end{figure*}

%% file: tex/04_method.tex
\section{Method\label{sec:method}}

\noindent{\textbf{{Overview.}}
As illustrated in~\cref{fig:pipeline}, our proposed framework for~\taskname, \modelname, contains four modules to cooperate with each other: a~\moduleAA, an~\moduleBB, a~\moduleCC, and a~\moduleDD. 

During training,~\modelname takes batched image-caption pairs as input. For each pair, the image is passed through the~\moduleAA~to be partitioned into several image segments, while the caption is first pre-processed to extract its linguistic structure as a text graph and then taken by the~\moduleBB~to ground textual entities in the graph onto the image segments. With the grounding results as pseudo labels, the~\moduleCC~learns similarity matrices between small image segments for further merging during inference, while the~\moduleDD~learns the prediction of object semantics and relation predicates.

During inference, for each input image, the image segments output from the~\moduleAA~are directly passed to the~\moduleCC~to be further merged according to the learned similarity matrices, and then fed to the~\moduleDD~to predict the object semantic labels and the relation predicate labels.

\subsection{Text Graph Preprocessing}
Following previous methods~\cite{Ye_2021_Linguistic, zhong_2021_SGGNLS, Li_2022_IntegratingOA} that leverage a rule-based language parser~\cite{Wu_2019_UnifiedVE} based on~\cite{Schuster_2015_GeneratingSP} to preprocess textual descriptions, in \modelname, we employ the same parser to extract linguistic structures from captions. Additionally, inspired by the success of open information extraction (OpenIE)~\cite{Angeli_2015_OpenIE} in natural language processing, we also employ an OpenIE system from Stanford CoreNLP~\cite{manning_2014_corenlp} for extraction as a supplement. After merging, for each caption, we obtain its linguistic structure represented in a text graph, where each node denotes an entity, and each directed edge denotes the relation between an entity pair.

\subsection{Region Grouper}
\label{sec:grouper}
With a scene image as input, the~\moduleAA~aims to merge the regions with similar object semantics into several segments and extract corresponding high-level features.

Our~\moduleAA~follows the hierarchical design of GroupViT~\cite{xu_2022_group}. Given an input image, the grouper first splits the image into $N$ non-overlapping patches as the initial image segments $\{\mathbf{s}_i^0\}_{i=1}^{N}$. These segments are then passed through $K$ grouping layers, where they are merged into larger, arbitrary-shaped segments progressively. Specifically, within each grouping layer $\mathbf{Grp}_k \ (k=1,2,\cdots,K)$, $H_k$ grouping centers $\{\mathbf{c}_i^k\}_{i=1}^{H_k}$ could be learned. The grouping operation is performed through an attention mechanism between the centers and the segments, merging $H_{k-1}$ input segments into $H_{k}$ larger ones, \ie, 
$$
\{\mathbf{s}_i^{k}\}_{i=1}^{H_{k}}
= 
\mathbf{Grp}_k(\{\mathbf{c}_i^k\}_{i=1}^{H_k},\{\mathbf{s}_i^{k-1}\}_{i=1}^{H_{k-1}}).
$$
Note that $H_0 = N$. After the hierarchical grouping, multiple groups of segments $\{\mathbf{s}_i^k\}_{i=1}^{H_k}$ at different grouping stages are obtained. More details about the design of $\{\mathbf{Grp}_k\}_{k=1}^K$ can be found in the appendix.

\subsection{Entity Grounder}
\label{sec:grounder}
Since the explicit region-entity links are not provided, bridging the textual description and the visual content automatically plays an important role in solving~\taskname.
Inspired by FILIP~\cite{yao_2022_filip}, in \modelname, we employ a similar fine-grained contrastive learning strategy to perform region-entity alignment.

For each grouping stage $k$, on image side, the grounder projects the segment group $\{\mathbf{s}_i^k\}_{i=1}^{H_k}$ into a new feature space $\mathcal{F}$ by a multi-layer perceptron (MLP) $\mathbf{Proj}_k^{I}$ to obtain segment embeddings $\{\mathbf{x}_i^{k}\}_{i=1}^{H_k}$.
On text side, the input caption is first tokenized into $M$ tokens $\{\mathbf{t}_i\}_{i=1}^M$, which are then processed by a Transformer~\cite{Vaswani_2017_AttentionIA} $\mathbf{Tfm}^{T}$ to propagate information between each other. A recurrent neural network (RNN) $\mathbf{Rnn}$ further 
merges the tokens corresponding to the same entity, encoding tokens into their associated weights one by one and utilizing weighted sum to merge the token features into a singular entity feature. Finally, these entity features are projected to the same feature space $\mathcal{F}$ by a MLP $\mathbf{Proj}^{T}$ to obtain entity embeddings $\{\mathbf{y}_i\}_{i=1}^{E}$, where $E$ denotes the number of entities in the caption.

With the segment embeddings and the entity embeddings in the shared feature space $\mathcal{F}$, we compute their token-wise similarities. Specifically, for the $i$-th segment, we compute its cosine similarities with all entities to obtain the token-wise similarity from the $i$-th segment to the caption $p_i^{k}$ via
$$
p_i^{k} = \max\limits_{1 \leq j \leq E} {\cos [\mathbf{x}_i^k,\mathbf{y}_j]},
$$
where $\cos [\cdot,\cdot]$ denotes the cosine similarity operation. 
Note that different from the original FILIP~\cite{yao_2022_filip}, in the scenario of region-entity alignment, some regions in the scene may not be described in the caption, while some entities in the caption may not exist in the scene. To tackle this problem, we propose to set a filtering threshold $\theta$, where pairs with similarity lower than $\theta$ will be considered in different semantics and filtered out.
The fine-grained similarity from the image to the caption $p^k$ can thus be computed via
$$
{p^k} = \frac{1}{\sum_{i=1}^{H_k}{\mathbf{1}_{p_i^{k} > \theta}}} \sum\limits_{i=1}^{H_k} {(p_i^{k}\cdot\mathbf{1}_{p_i^{k} > \theta}}).
$$
Similarly, we can also compute the token-wise similarity from the $j$-th entity to the image $q_j^k$ via
$$
q_j^{k} = \max\limits_{1 \leq i \leq H_k} {\cos [\mathbf{x}_i^k,\mathbf{y}_j]},
$$
and the fine-grained similarity from the caption to the image $q^k$ via
$$
q^{k} = \frac{1}{\sum_{j=1}^{E}{\mathbf{1}_{q_j^{k} > \theta}}} \sum\limits_{j=1}^{E} {(q_j^{k}\cdot\mathbf{1}_{q_j^{k} > \theta}}).
$$

Denoting the training batch with batch size $B$ as $\{(I_i, T_i)\}_{i=1}^B$, the fine-grained similarity from the image $I_i$ to the caption $T_j$ as $p^{k, i \rightarrow j}$ and from the caption $T_j$ to the image $I_i$ as $q^{k, j \rightarrow i}$, the image-to-text fine-grained contrastive loss $\mathcal{L}_{fine}^{k,I \rightarrow T}$ and the text-to-image fine-grained contrastive loss $\mathcal{L}_{fine}^{k,T \rightarrow I}$ can then be formulated as 
$$
\mathcal{L}_{fine}^{k,I \rightarrow T} = -\frac{1}{B}\sum\limits_{i=1}^B \frac{\exp{(p^{k, i \rightarrow i} / {\tau})}}{\sum\limits_{j=1}^B {\exp{(p^{k, i \rightarrow j} / {\tau})}}},
$$
$$
\mathcal{L}_{fine}^{k,T \rightarrow I} = -\frac{1}{B}\sum\limits_{i=1}^B \frac{\exp{(q^{k, i \rightarrow i} / {\tau})}}{\sum\limits_{j=1}^B {\exp{(q^{k, i \rightarrow j} / {\tau})}}},
$$
where $\tau$ is a learnable temperature. The total fine-grained contrastive loss is 
$$
\mathcal{L}_{fine}^k = \frac{1}{2}(\mathcal{L}_{fine}^{k,I \rightarrow T} + \mathcal{L}_{fine}^{k,T \rightarrow I}).
$$

By minimizing $\mathcal{L}_{fine}^k$ at all grouping stages during training, our framework could reach a meaningful fine-grained alignment automatically, \ie, for the $i$-th segment $\mathbf{s}_i^k$, the $l_i^k$-th entity satisfying
$$
l_i^k = \underset{1 \leq j \leq E}{\arg\max\limits} {\ \cos [\mathbf{x}_i^k,\mathbf{y}_j]}
$$
tends to have a similar semantics with $\mathbf{s}_i^k$. We thus obtain $\{l_i^k\}_{i=1}^{H_k}$ as the grounding results for the image segments $\{\mathbf{s}_i^k\}_{i=1}^{H_k}$.
A further explanation of the automatic meaningful alignment can be found in the appendix.

\subsection{Segment Merger}
\label{sec:merger}
To improve the ability of location, we propose to leverage the entity-grounding results as explicit supervision to learn a group of similarity matrices between image segments for small segments merging during inference, compared with~\cite{xu_2022_group} that learns the merging fully implicitly.

For each grouping stage $k$, we compute the cosine similarity between each pair of image segments, which is then linearly re-scaled into $[0,1]$ to formulate a similarity matrix $\mathbf{Sim}_{k} \in [0,1]^{H_k \times H_k}$, where
$$
\mathbf{Sim}_{k} [i,j] = \frac{1}{2}(\cos [\mathbf{x}_i^k, \mathbf{x}_j^k] + 1).
$$

 We further leverage $\{l_i^k\}_{i=1}^{H_k}$ as pseudo labels to build a pseudo target matrix $\mathbf{Sim}_k^{target}\in \{0,1\}^{H_k \times H_k}$, where
$$
\mathbf{Sim}_{k}^{target} [i,j] = \left\{
\begin{aligned}
    1, & \ \mathbf{if} \ l_i^k = l_j^k \land \cos[\mathbf{x}^k_i, \mathbf{y}_{l_i^k}] > \theta \\
    & \ \land \cos[\mathbf{x}^k_j, \mathbf{y}_{l_j^k}] > \theta, \\
    0, & \ \mathbf{otherwise}.
\end{aligned}
\right.
$$

The similarity loss for the stage $k$ is then formulated as 
$$
\mathcal{L}_{sim}^{k} = \frac{1}{H_k^2} \|\mathbf{Sim}_k - \mathbf{Sim}_k^{target}\|_F^2.
$$

\subsection{Label Generator}
\label{sec:generator}
In addressing the challenge of no pre-defined concept sets, the previous work~\cite{zhong_2021_SGGNLS} proposes to build a large vocabulary for learning during training and use WordNet~\cite{miller_1992_wordnet} to correlate predictions within this vocabulary to the target concepts during inference. 
However, there are two limitations to the previous method. 
Firstly, compared with the extensive object semantics and relation predicates contained in textual descriptions, despite the large vocabulary established, it is inevitable that some classes will be overlooked. 
 Secondly, leveraging WordNet to match vocabulary predictions to targets is not accurate and robust enough, for WordNet may only reach a coarse matching with multiple target concepts. This imprecision is particularly pronounced for relation predicates relative to object semantics.

Given these limitations, we introduce a novel approach in~\modelname. Instead of approaching label prediction of objects and relations as a traditional classification problem, we reformulate it as an auto-regressive generation problem, which eliminates the necessity for pre-defined concept sets.

Compared with a vanilla RNN, we employ a pre-trained vision language model BLIP~\cite{li_2022_blip}to leverage the pre-learned common sense. BLIP can take an image as input and output a caption to describe the image. In~\modelname, we borrowed the pre-trained decoder module from BLIP to perform the generation of object and relation labels. 

During training, the~\moduleDD~takes the caption-parsed text graph, the segment features from the~\moduleAA, and the grounding results $\{l_i^k\}_{i=1}^{H_k}$ from the~\moduleBB~as input. It filters out the segments with token-wise similarity lower than the threshold $\theta$, merges the segments mapped to the same entity, and queries the corresponding image masks from the~\moduleAA. Then, $E_k$ image masks $\{\mathbf{m}_i^k\}_{i=1}^{E_k}$ with their pseudo entity labels $\{b_i^k\}_{i=1}^{E_k}$ can be obtained, where each $b_i^k$ is one entity in the text graph. $E_k \leq E$ because some textual entities may not exist in the image.

\noindent{\textbf{{Prompt-embedding-based technique (PET).}}
To better incorporate common sense from the vision language model, we further design a novel PET for label generation. For object prediction, the decoder takes the segment features and the image mask $\mathbf{m}_i^k$, using a prompt
$$
\mathbf{
a \  photo \  of \  [ENT]
}
$$
to guide the object generation, where the $\mathbf{[ENT]}$ token is expected to be the pseudo label $b_i^k$. For relation prediction, the decoder takes the segment features and an image mask pair $(\mathbf{m}_i^k, \mathbf{m}_j^k)$ as input, using a prompt
$$
\mathbf{
a \  photo \  of \  [SUB]  \ and  \ [OBJ] 
}
$$
$$
\mathbf{
what \  is \  their \  relation \  [REL]
}
$$
to guide the relation generation, where the $\mathbf{[SUB]}$ and $\mathbf{[OBJ]}$ tokens are embedded by the pseudo labels $b_i^k$ and $b_j^k$, and the $\mathbf{[REL]}$ token is expected to be the relation predicate between $(b_i^k,b_j^k)$ with $b_i^k$ as subject and $b_j^k$ as object in the text graph. 
To enhance relation generation, we further design three learnable positional embeddings $\mathbf{f}_{sub}$, $\mathbf{f}_{obj}$, $\mathbf{f}_{region}$ for indicating the different regions in the segment features.
Two cross-entropy losses $\mathcal{L}_{ent}^k, \mathcal{L}_{rel}^k$ are used to supervise the generation of the $\mathbf{[ENT]}$ and $\mathbf{[REL]}$ tokens, maximizing the likelihood of the target label strings, respectively. 
More details about PET can be found in the appendix.

\subsection{Inference}
\label{sec:inference}
During inference, the target concepts of object semantics $\mathcal{C}_o$ and relation predicates $\mathcal{C}_r$ are known.
With an image $I$ as input and an inference stage index $k_{inf}$ specified, the~\moduleAA~first partitions $I$ into several candidate segments $\{\mathbf{s}_i^{k_{inf}}\}_{i=1}^{H_{k_{inf}}}$, which are then passed through the~\moduleCC~to obtain the similarity matrix $\mathbf{Sim}_{k_{inf}}$. 
We formulate the segment merging as a spectral clustering problem and perform the graph cut~\cite{shi_2000_ncut} on $\mathbf{Sim}_{l_{inf}}$ for clustering. To improve the accuracy, we employ a matrix recovery method~\cite{liu_2012_robust} to reduce the noise in $\mathbf{Sim}_{l_{inf}}$. In this step, the segments with similar semantics tend to be merged into the same cluster. 
For each cluster and each pair of clusters, the~\moduleDD~use a similar PET to generate the object semantics and the relation predicates. For every label within sets $\mathcal{C}_o$ and $\mathcal{C}_r$, the~\moduleDD~computes its generation probability. Subsequently, these probabilities are used to rank the concepts, selecting the most probable as the final prediction.
Note that between object and relation prediction, to convert semantic segmentation into instance segmentation, we identify each connected component in the semantic segmentation to be an instance, for simplicity. 
More details about inference can be found in the appendix.

%% file: tex/05_experiments.tex
\section{Experiments and Results\label{sec:experiments}}

\input{table/01_main_result.tex}

\noindent{\textbf{Datasets.}}
We train our model with the COCO Caption dataset~\cite{chen_2015_caption}, which involves 123,287 images with each labeled by 5 independent human-generated captions. Following the 2017 split, we use 118,287 images with their captions for training.
We evaluate models with the Panoptic Scene Graph dataset ~\cite{yang_2022_psg} for its pixel-wise labeling as well as its high-quality object and relation annotation.
We further merge the object semantics with ambiguities. After merging, 127 object semantics and 56 relation predicates are finally obtained for evaluation.
More details about the datasets can be found in the appendix.

\noindent{\textbf{Evaluation Protocol and Metrics.}}
Following all previous works in scene graph generation, we evaluate the quality of a generated scene graph by viewing it as a set of subject-predicate-object triplets.
We evaluate models on two tasks: Visual Phrase Detection (\textbf{PhrDet}) and Scene Graph Detection (\textbf{SGDet}). \textbf{PhrDet} aims to detect the whole phrase of subject-predicate-object with a union location of subject and object. It is considered to be correct if the phrase labels are all correct and the union location matches the ground truth with the intersection over union (IoU) greater than 0.5. \textbf{SGDet} further requires a more accurate location, \ie, the location of subject and object should match the ground truth with IoU greater than 0.5 respectively.

We use No-Graph-Constraint-X Recall@K (\textbf{NXR@K}, \%) to measure the ability of generation. Recall@K computes the recall between the top-k generated triplets with the ground truth.
No-Graph-Constraint-X indicates that at most X predicate labels could be predicted for each subject-object pair.
Since some predicates defined in~\cite{yang_2022_psg} are not exclusive, such as \textit{on} and \textit{sitting on}, \textbf{NXR@K} could be a more reasonable metric compared with Recall@K.

\noindent{\textbf{Baselines.}}
We consider several baselines for \taskname in the following experiments. Firstly, we design four baselines that strictly follow the constraints of \taskname, where objects are located by bbox proposals generated by selective search~\cite{uijlings_2013_selective}:
\begin{itemize}[align=right,itemindent=0em,labelsep=2pt,labelwidth=1em,leftmargin=*,itemsep=0em] 
\item \textbf{Random} is the most naive baseline where all object semantics and relation predicates are randomly predicted.

\item \textbf{Prior} augments \textbf{Random} by performing label prediction based on the statistical priors in the training set.

\item \textbf{MIL} performs the alignment between proposals and textual entities by multiple instance learning~\cite{maron_1997_mil}. Similar to~\cite{zhong_2021_SGGNLS}, it formulates the object label prediction as a classification problem in
a large pre-built vocabulary, with WordNet~\cite{miller_1992_wordnet} employed during inference. The relation labels are predicted with statistical priors, similar to \textbf{Prior}.

\item \textbf{SGCLIP} employs the pre-trained CLIP~\cite{radford_2021_clip} to predict both object semantic labels and relation predicate labels.

Secondly, to further benchmark the performance of our framework, we set two additional baselines based on~\cite{zhong_2021_SGGNLS} by gradually removing the constraints of~\taskname:

\item \textbf{SGGNLS-o}~\cite{zhong_2021_SGGNLS} extracts proposals with a detector~\cite{ren_2015_fasterrcnn} pre-trained on OpenImage~\cite{Kuznetsova_2018_OpenImage}. 
It formulates the object and relation label prediction as a classification problem within a large pre-built vocabulary, with WordNet~\cite{miller_1992_wordnet} employed during inference.

\item \textbf{SGGNLS-c}~\cite{zhong_2021_SGGNLS} uses the same proposals as \textbf{SGGNLS-o}. In~\textbf{SGGNLS-c}, the target concept sets for inference are known during training. It formulates the label prediction as a classification problem in these target concept sets.
\end{itemize}

More design details can be found in the appendix.

\noindent{\textbf{Implementation Details.}} Following GroupViT~\cite{xu_2022_group}, we set $K=2$, $H_1=64$, and $H_2=8$ for our~\moduleAA. We leverage general pre-trained models for weight initialization. We employ the pre-trained GroupViT for the~\moduleAA~as well as $\mathbf{Tfm}^T$ in the~\moduleBB, and the pre-trained BLIP~\cite{li_2022_blip} decoder for the~\moduleDD.
During training, $\mathbf{Tfm}^T$ and the~\moduleDD~are frozen.
During inference, we set $k_{inf}=1$. More implementation details can be found in the appendix.

\input{figure/03_visualization.tex}

\subsection{Main Results on \taskname}

\noindent{\textbf{{Quantitative Results.}}
Our quantitative results on \taskname are shown in~\cref{tab:main-result}. To make a fair comparison with bbox-based scene graphs generated by baselines, we evaluate our generated PSGs in both mask and bbox mode. For the latter, all masks in both prediction and ground truth are converted into bboxes (\ie, the mask area's enclosing rectangle) for evaluation, resulting in an easier setting than the former. The results show that our framework (\textbf{Ours}) significantly outperforms all the baselines under the same constraints on both PhrDet (14.37 vs. 3.71 N5R100) and SGDet (5.48 vs. 2.7 N5R100).
Our method also shows better results compared with \textbf{SGGNLS-o} on all metrics and all tasks (on PhrDet, 14.37 vs. 7.93 N5R100; on SGDet, 5.48 vs. 5.02 N5R100) although \textbf{SGGNLS-o} utilizes location priors by leveraging a pre-trained detector. The results demonstrate that our framework is more effective for learning a good panoptic structured scene understanding.

\noindent{\textbf{Qualitative Results.}}
We provide typical qualitative results in~\cref{fig:visualization} to further show our framework's effectiveness.
Compared with \textbf{SGGNLS-o}, our framework has the following advantages. First, our framework is able to provide fine-grained semantic labels to each pixel in the image to reach a panoptic understanding, while \textbf{SGGNLS-o} can only provide sparse bboxes produced by the pre-trained detector. Note that categories with irregular shapes (\eg, trees in~\cref{fig:visualization}) are hard to be labeled precisely by bboxes. Second, compared with \textbf{SGGNLS-o}, our framework can generate more comprehensive object semantics and relation predicates, such as \textit{``dry grass field"} and ``\textit{land at}" in~\cref{fig:visualization}, showing the open-vocabulary potential of our framework.
More qualitative results can be found in the appendix.

\subsection{OOD Robustness Analysis}
We further analyze another key advantage of our framework, \ie, the robustness in OOD cases}.
Since \textbf{SGGNLS-c} and \textbf{SGGNLS-o} both rely on a pre-trained detector to locate objects, their performance highly depends on whether object semantics in the scene are covered by the detector. 

\input{table/02_ood.tex}

Based on the object semantics~\cite{Kuznetsova_2018_OpenImage} covered by the detector, we split the ground truth triplets into an in-distribution (ID) set and an OOD set. For triplets within the ID set, both the subject and object semantics are covered, while for triplets in the OOD set, at least one of the semantics is not covered.
As shown in~\cref{tab:ood}, both \textbf{SGGNLS-c} and \textbf{SGGNLS-o} suffer a significant performance drop from the ID set to the OOD set. On the OOD set, the triplets can hardly be retrieved. However, our framework, with the ability of location learned from purely text descriptions, can reach similar performance on both sets, which demonstrates the OOD robustness of our framework for PSG generation.

\input{table/03_ablation_merger.tex}

\subsection{Ablation Studies}

We conduct additional ablation studies to evaluate the effectiveness of our design choices. 
For all following experiments, we report N3R100 and N5R100 evaluated in bbox mode for simplicity.
We answer the following questions.
\textbf{Q1}: Does the explicit learning of merging in the~\moduleCC~helps provide better image segments?
\textbf{Q2}: Is the generation-based label prediction better than the classification-based prediction?
\textbf{Q3}: Does the pre-learned common sense from the pre-trained BLIP~\cite{li_2022_blip} helps with the label prediction?
\textbf{Q4}: Does the PET helps incorporate the pre-learned common sense for label prediction?

In~\cref{tab:abla-merger}, we compare different strategies of image segment merging during inference. Row 1\&2 denote that the $H_1=64$ segments from the first grouping stage are used for further merging, while row 3\&4 denote that the $H_2=8$ segments from the second stage are used. 
The results show that applying the graph cut to merge the segments from the first stage could reach the best performance, corroborating that compared with the fully implicit learning of merging, the explicit learning of merging can provide better segments (row 2 vs 3, answering \textbf{Q1}).

\input{table/04_ablation_generator.tex}

In~\cref{tab:abla-generator}, we compare different designs of the~\moduleDD. Keeping the other modules the same, we change the~\moduleDD~(row 4) into three different designs, \ie, classification within a large pre-built vocabulary followed by WordNet~\cite{miller_1992_wordnet} for target matching (row 1), generation with a vanilla RNN (row 2), generation with the BLIP decoder but without the PET (row 3).
The results show that with the constraint of no pre-defined concept sets, compared with formulating the label prediction into a classification problem, formulating it into a generation problem is a better choice (row 1 vs 2\&4, answering \textbf{Q2}). By employing the pre-trained BLIP for leveraging the pre-learned common sense, the prediction could be further boosted (row 2 vs 4, answering \textbf{Q3}). And the PET is very important for incorporating the common sense from the pre-trained model (row 3 vs 4, answering \textbf{Q4}).

More ablation studies for the design evaluation can be found in the appendix.

\subsection{Application on TSSS}
\input{table/05_sem_seg.tex}
As a side product, we observe that our~\moduleBB~and~\moduleCC~can also enhance TSSS. 
Based on the original GroupViT~\cite{xu_2022_group}, we replace the multi-label contrastive loss with our~\moduleBB~and~\moduleCC.
Then we finetune the model on the COCO Caption dataset~\cite{chen_2015_caption}.
As shown in~\cref{tab:semseg}, compared with GroupViT directly finetuned on~\cite{chen_2015_caption}, the explicit learning of merging in our modules can boost the model with an absolute 2.15\% improvement of mean Intersection over Union (mIoU, \%) on COCO~\cite{Lin_2014_COCO}, which demonstrates the effectiveness of our proposed modules on better object location.

\subsection{Discussion}
\noindent{\textbf{Failure Case Analysis.}}
Despite the impressive performance of~\modelname, there are still challenges to address. Upon analyzing the failure cases for PSG generation, we identify three specific limitations of~\modelname that contribute to these failures. 
a) The strategy we use to convert semantic segmentation into instance segmentation is not entirely effective. For simplicity, in~\modelname, we identify each connected component in the semantic segmentation to be an individual object instance. However, this strategy may fail when instances overlap or are occluded, resulting in either an underestimation or an overestimation of instances. 
b) Our framework faces difficulty in locating small objects in the scene due to limitations in resolution and the grouping strategy for location.
c) The relation prediction of our framework requires enhancement, as it is not adequately conditioned on the image. While the~\moduleDD~uses both image features and predicted object semantics to determine the relation, it sometimes seems to lean heavily on the object semantics, potentially neglecting the actual image content. Examples of failure cases for each of these limitations can be found in the appendix.

\noindent{\textbf{Model Diagnosis.}} 
For a clearer understanding of the efficacy of our framework, we conduct a model diagnosis to answer the following question: why does our framework only achieve semantic segmentation through learning, rather than panoptic segmentation (and thus requires further segmentation conversion to obtain instance segmentation)?

\input{figure/05_caption_granularity.tex}

In~\cref{fig:caption_granularity}, we use two captions in different granularity to execute region-entity alignment. It shows that our framework has the capability to assign distinct masks to individual instances. However, the nature of caption data, where captions often merge objects of the same semantics in plural form, limits our framework from differentiating instances. It is the weak supervision provided by the caption data that constrains our framework.

More diagnoses can be found in the appendix.

\noindent{\textbf{Future Directions.}} In response to the limitations discussed, we outline several potential directions for enhancing our framework: a) a refined and sophisticated strategy for segmentation conversion; b) increasing the input resolution, though this may introduce greater computational demands; c) a more suitable image-conditioned reasoning mechanism for relation prediction; d) a superior image-caption-pair dataset with more detailed granularity in captions to achieve panoptic segmentation through learning.

%% file: table/01_main_result.tex
\begin{table*}[t]
\resizebox{\linewidth}{!}{
\centering
\footnotesize
\renewcommand\arraystretch{1.2} 
\begin{tabular}{ccc|c|cccc|cccc}
\shline
\multicolumn{3}{c|}{Method} & \multirow{2}{*}{Mode} & \multicolumn{4}{c|}{PhrDet} & \multicolumn{4}{c}{SGDet} \\ 
\cline{1-3} 
\multicolumn{1}{c|}{Model} & \multicolumn{1}{c|}{Proposal} & Target &  & N3R50 & N3R100 & N5R50 & N5R100 & N3R50 & N3R100 & N5R50 & N5R100 \\ 
\shline
\multicolumn{1}{c|}{SGGNLS-c} & \multicolumn{1}{c|}{Detector} & \ding{52} & bbox & 9.69 & 11.45 & 10.24 & 12.22 & 6.76 & 7.81 & 7.2 & 8.65 \\ 
\hline\hline
\multicolumn{1}{c|}{Random} & \multicolumn{1}{c|}{\multirow{4}{*}{\makecell{Selective \\ Search}}} & \ding{56} & bbox & 0.02 & 0.03 & 0.02 & 0.03 & 0.01 & 0.02 & 0.02 & 0.03 \\ 
\multicolumn{1}{c|}{Prior} & \multicolumn{1}{c|}{} & \ding{56} & bbox & 0.04 & 0.07 & 0.05 & 0.07 & 0.03 & 0.06 & 0.05 & 0.07 \\ 
\multicolumn{1}{c|}{MIL} & \multicolumn{1}{c|}{} & \ding{56} & bbox & 1.97 & 2.18 & 2.04 & 2.61 & 1.2 & 1.35 & 1.56 & 1.97 \\ 
\multicolumn{1}{c|}{SGCLIP} & \multicolumn{1}{c|}{} & \ding{56} & bbox & 3.02 & 3.45 & 3.38 & 3.71 & 2.13 & 2.3 & 2.39 & 2.7 \\ 
\hline\hline
\multicolumn{1}{c|}{SGGNLS-o} & \multicolumn{1}{c|}{Detector} & \ding{56} & bbox & 6.2 & 6.79 & 6.92 & 7.93 & 3.96 & 4.21 & 4.53 & 5.02 \\ 
\hline\hline
\multicolumn{1}{c|}{Ours} & \multicolumn{1}{c|}{\textendash} & \ding{56} & mask & 8.28 & 9.16 & 9.06 & 10.51 & 3.32 & 3.63 & 3.71 & 4.18 \\ 
\multicolumn{1}{c|}{Ours} & \multicolumn{1}{c|}{\textendash} & \ding{56} & bbox & \textbf{11.37} & \textbf{12.74} & \textbf{12.24} & \textbf{14.37} & \textbf{4.29} & \textbf{4.77} & \textbf{4.82} & \textbf{5.48} \\ 
\shline
\end{tabular}}
\caption{\textbf{Quantitative Comparison of Different Methods on \taskname.} `Proposal' indicates how the method obtains bbox proposals. `Target' indicates whether the concept sets for inference are known during training. `Mode' indicates the mode used for evaluation.}
\label{tab:main-result}
\end{table*}

%% file: figure/03_visualization.tex
\begin{figure*}[t]
  \centering
  \includegraphics[width=1\linewidth]{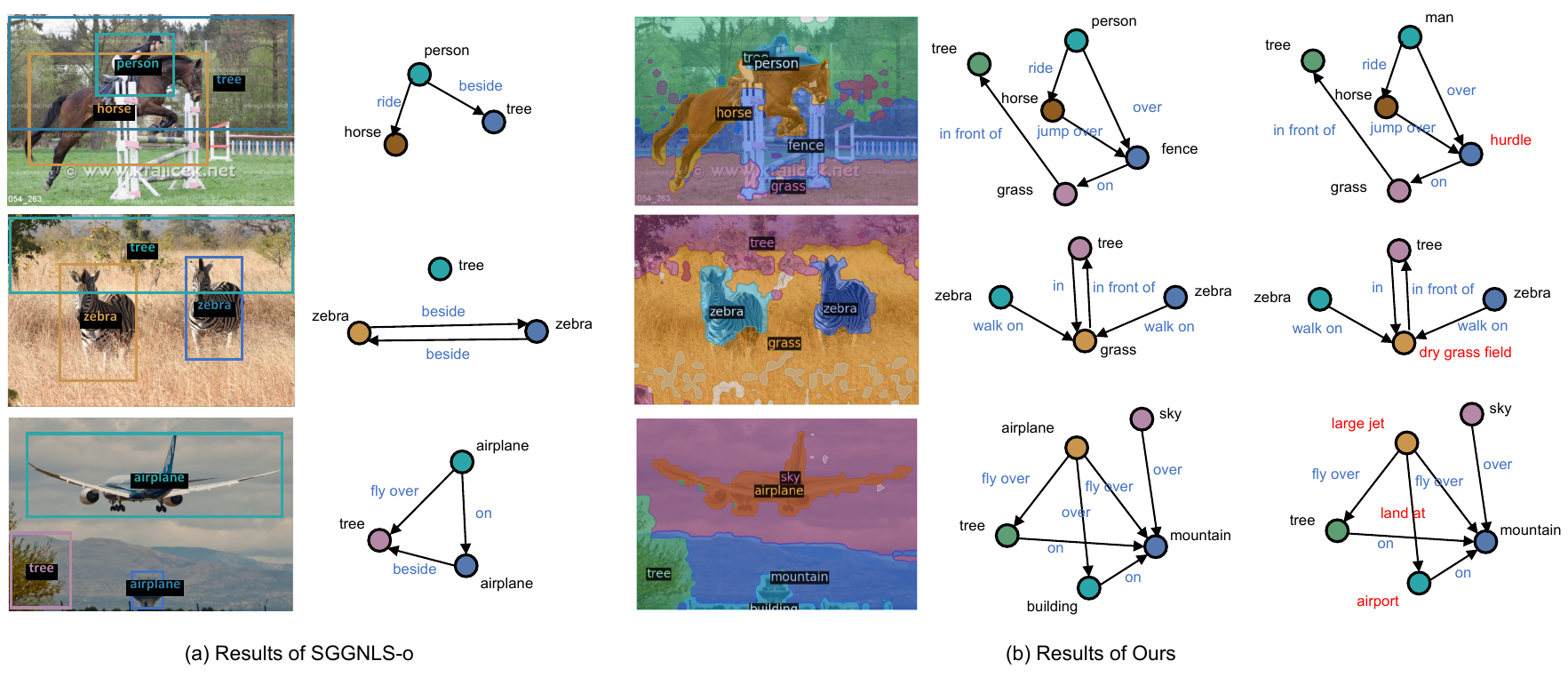}
   \caption{\textbf{Qualitative Comparison between SGGNLS-o (a) and Ours (b).} For each method, the results of object location are shown on the left, while the results of scene graph generation are shown on the right. For \textbf{Ours}, scene graphs predicted within the given concept sets are provided in the middle column, and scene graphs directly predicted through the auto-regressive generation (\ie, an open-vocabulary manner) in the~\moduleDD~are additionally provided in the right column.}
   \label{fig:visualization}
\end{figure*}

%% file: table/02_ood.tex
\begin{table}[t]
\resizebox{\linewidth}{!}{
\setlength{\tabcolsep}{4pt}
\renewcommand{\arraystretch}{1.1}
\begin{tabular}{c|c|c|c|cc|cc}
\shline
\multicolumn{1}{l|}{\multirow{2}{*}{Set}} & \multirow{2}{*}{Model} & \multirow{2}{*}{Target} & \multirow{2}{*}{Mode} & \multicolumn{2}{c|}{PhrDet} & \multicolumn{2}{c}{SGDet} \\ 
\multicolumn{1}{l|}{} &  &  &  & N3R100 & N5R100 & N3R100 & N5R100 \\ 
\shline
\multirow{4}{*}{ID} & SGGNLS-c & \ding{52} & bbox & \textbf{16.76} & \textbf{18.48} & \textbf{10.45} & \textbf{11.86} \\
 & SGGNLS-o & \ding{56} & bbox & 11.55 & 13.64 & 7.13 & 8.47 \\
  \cline{2-8}
 & Ours & \ding{56} & mask & 9.27 & 10.45 & 3.28 & 3.76 \\ 
 & Ours & \ding{56} & bbox & 13.35 & 14.82 & 4.63 & 5.36 \\
 \hline
  \hline
\multirow{4}{*}{OOD} & SGGNLS-c & \ding{52} & bbox & 0 & 0 & 0 & 0 \\
 & SGGNLS-o & \ding{56} & bbox & 0.05 & 0.06 & 0 & 0 \\
  \cline{2-8}
& Ours & \ding{56} & mask & 8.47 & 9.76 & 4.07 & 4.51 \\
 & Ours & \ding{56} & bbox & \textbf{10.18} & \textbf{11.69} & \textbf{5.23} & \textbf{5.72} \\
 \shline
\end{tabular}}
\caption{\textbf{Analysis on OOD Robustness.} `Set' indicates the triplet set used for evaluation.}
\label{tab:ood}
\end{table}

%% file: table/03_ablation_merger.tex
\begin{table}[t]
\resizebox{\linewidth}{!}{
\setlength{\tabcolsep}{8pt}
\begin{tabular}{l|c|c|cc|cc}
\shline
\multirow{2}{*}{Stage} & \multirow{2}{*}{\#Seg} & \multirow{2}{*}{Cut} & \multicolumn{2}{c|}{PhrDet} & \multicolumn{2}{c}{SGDet} \\
 &  &  & N3R100 & N5R100 & N3R100 & N5R100 \\ 
 \hline
 1 & 64 & \ding{56} & 10.73 & 11.39 & 3.18 & 3.51 \\
 1 & 64 & \ding{52} & \textbf{12.74} & \textbf{14.37} & \textbf{4.77} & \textbf{5.48} \\
2 & 8 & \ding{56} & 9.24 & 11.03 & 3.53 & 4.35 \\
2 & 8 & \ding{52} & 6.78 & 8.45 & 2.46 & 3.21 \\
\shline
\end{tabular}}
\caption{\textbf{Ablation Study on the Segment Merger.} `Stage' indicates the grouping stage where image segments used for merging are from. `\#Seg' indicates the number of image segments. `Cut' indicates whether the graph-cut-based segment merging is applied.}
\label{tab:abla-merger}
\end{table}

%% file: table/04_ablation_generator.tex
\begin{table}[t]
\resizebox{\linewidth}{!}{
\begin{tabular}{c|c|cc|cc}
\shline
\multirow{2}{*}{Label Prediction} & \multirow{2}{*}{Model} & \multicolumn{2}{c|}{PhrDet} & \multicolumn{2}{c}{SGDet} \\
 &  & N3R100 & N5R100 & N3R100 & N5R100 \\ \hline
Cls + WordNet & - & 8.82 & 9.36 & 2.36 & 2.72 \\
Gen & RNN & 9.12 & 10.44 & 2.65 & 3.07 \\
Gen w/o PET & BLIP~\cite{li_2022_blip} & 2.33 & 2.58 & 0.45 & 0.6 \\
Gen w/ PET & BLIP~\cite{li_2022_blip} & \textbf{12.64} & \textbf{14.28} & \textbf{4.77} & \textbf{5.49} \\
\shline
\end{tabular}}
\caption{\textbf{Ablation Study on the Label Generator.} `Cls' indicates classification. `Gen' indicates generation.}
\label{tab:abla-generator}
\end{table}

%% file: table/05_sem_seg.tex
\begin{wraptable}{r}{4.1cm}
\centering
\small
\setlength{\tabcolsep}{6pt}
\begin{tabular}{cc}
\hline
\multicolumn{1}{c|}{Method} & mIoU \\ 
\hline
\multicolumn{1}{c|}{GroupViT~\cite{xu_2022_group}} & 24.28 \\ 
\multicolumn{1}{c|}{GroupViT$^{\dag}$~\cite{xu_2022_group}} & 24.72 \\ 
\multicolumn{1}{c|}{Ours} & \textbf{26.87} \\ 
\hline
\end{tabular}
\caption{\textbf{Results on TSSS.} $^\dag$ indicates finetuned.}
\label{tab:semseg}
\end{wraptable}

%% file: figure/05_caption_granularity.tex
\begin{figure}[h]
  \centering
  \includegraphics[width=1\linewidth]{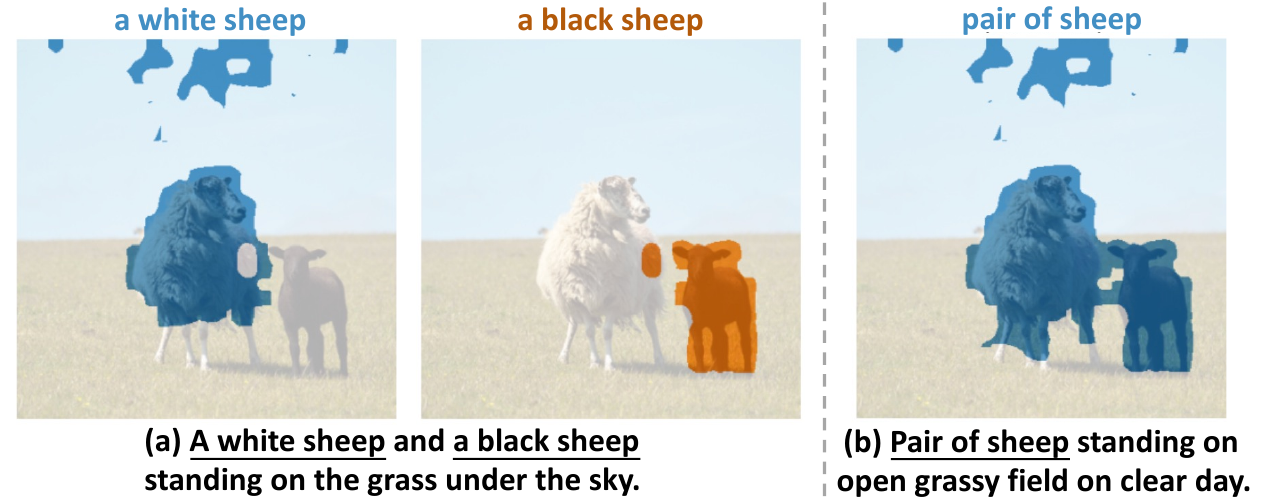}
   \caption{\textbf{Region-Entity Alignment Results of Captions in Different Granularity.} Two captions in different granularity are used to execute region-entity alignment with the same image, with (a) one describing the two sheep individually while (b) the other merges them in plural form. 
   }
   \label{fig:caption_granularity}
\end{figure}

%% file: tex/06_conclusion.tex
\section{Conclusion\label{sec:conclusion}}
We take the first step towards the novel problem \taskname, aiming to learn PSG generation purely from language.
To tackle this challenging problem, we propose a new modularized framework~\modelname with several novel techniques, which significantly outperforms the baselines and achieves strong OOD robustness. This paves the path to a more comprehensive and generalizable panoptic structured scene understanding.
There are still bottlenecks in~\modelname to be explored in future work, including a) a more sophisticated strategy for segmentation conversion; b) a more suitable image-conditioned reasoning mechanism for relation prediction; c) a superior image-caption-pair dataset for panoptic segmentation through learning.

\paragraph{Acknowledgements.} 
This work was supported by the DSO grant DSOCO21072, IBM, and gift funding from MERL, Cisco, and Amazon. We would also like to thank the computation support from AiMOS, a server cluster for the IBM Research AI Hardware Center.

%% file: tex/ap_01_method.tex
\section{More Details of TextPSG Framework}
\label{sec:supp-framework}

\subsection{More Details of Region Grouper}

The~\moduleAA~follows the design of GroupViT~\cite{xu_2022_group}. The input scene image $I$ is first split into $N$ non-overlapping patches and projected to be initial image segments $\{\mathbf{s}_i^0\}_{i=1}^N$, which are then passed through $K$ grouping layers $\{\mathbf{Grp}_k\}_{k=1}^K$ to be merged progressively. Each grouping layer $\mathbf{Grp}_k$ consists of $H_k$ learnable grouping centers $\{\mathbf{c}_i^k\}_{i=1}^{H_k}$, a Transformer~\cite{Vaswani_2017_AttentionIA}-based block $\mathbf{Tfm}_k^I$ for communication between the centers $\{\mathbf{c}_i^k\}_{i=1}^{H_k}$ and the segments $\{\mathbf{s}_i^{k-1}\}_{i=1}^{H_{k-1}}$, and an attention-based block $\mathbf{Att}_k$ for assigning the segments to different centers and merging the segments corresponding to the same center into $\{\mathbf{s}_i^{k}\}_{i=1}^{H_{k}}$. Within $\mathbf{Grp}_k$, the grouping is performed as
$$
\{\mathbf{s}_i^{k}\}_{i=1}^{H_{k}}
= 
\mathbf{Att}_k(\mathbf{Tfm}_k^I(\{\mathbf{c}_i^k\}_{i=1}^{H_k},\{\mathbf{s}_i^{k-1}\}_{i=1}^{H_{k-1}})).
$$
Note that $H_0 = N$. Especially, the updated image segments $\{\hat{\mathbf{s}}_i^0\}_{i=1}^{H_0}$ from the communication block $\mathbf{Tfm}_1^I$ in the first grouping layer $\mathbf{Grp}_1$ will be further used by the~\moduleDD~for the label prediction, as introduced in the following.

\subsection{More Details of Entity Grounder}
In the~\moduleBB, meaningful region-entity alignment can be reached automatically during training, serving as pseudo labels for the learning of the~\moduleCC~and the~\moduleDD. Here we provide a further explanation of the automatic meaningful alignment. 

In the~\moduleBB, the total fine-grained contrastive loss $\mathcal{L}_{fine}^k$ consists of two symmetry components $\mathcal{L}_{fine}^{k, I \rightarrow T}$ and $\mathcal{L}_{fine}^{k, T \rightarrow I}$. 
Minimizing $\mathcal{L}_{fine}^k$ equals to minimizing $\mathcal{L}_{fine}^{k,I \rightarrow T}$ and $\mathcal{L}_{fine}^{k,T \rightarrow I}$ simultaneously. 

Here we take $\mathcal{L}_{fine}^{k,I \rightarrow T}$ as an example while the other remains the same. 
In each batch, we assume that for each region in each image, there is at most one corresponding entity in the corresponding caption, while all the other entities in the caption and all entities in the other captions are mismatched with the region.

To minimize $\mathcal{L}_{fine}^{k,I \rightarrow T}$, for each image $I_i$ in the batch, the model needs to maximize $p^{k,i \rightarrow i}$ and minimize all other $p^{k,i \rightarrow j}$ where $j \neq i$. 

To minimize $p^{k,i \rightarrow j}$, with $p^{k,i \rightarrow j}$ denoting the mean value of $p^{k,i \rightarrow j}_l$ and $l$ for the index of the region in the image $I_i$, the model needs to minimize all $p^{k,i \rightarrow j}_l$. Since $p^{k,i \rightarrow j}_l$ denotes the max cosine similarity between the $l$-th region and all entities in $T_j$, minimizing $p^{k,i \rightarrow j}_l$ equals pushing the $l$-th region and all entities in $T_j$ apart in the shared feature space. 

To maximize $p^{k,i \rightarrow i}$, the model needs to maximize all $p^{k,i \rightarrow i}_l$. A global maximum is that the $l$-th region is close to the corresponding entity in $T_i$ and far from all the other entities in the shared feature space. 

By minimizing $p^{k,i \rightarrow j}$ and maximizing $p^{k,i \rightarrow i}$ at the same time, the model tends to pull similar region-entity pairs to be closer and push dissimilar pairs apart in the shared feature space, thus reaching a meaningful region-entity alignment automatically.

\subsection{More Details of Label Generator}
Here we provide more details about the prompt-embedding-based technique (PET) used in the~\moduleDD.

To predict the object semantics, for each image mask $\mathbf{m}_i^k$, the~\moduleDD~ takes the updated image tokens $\{\hat{\mathbf{p}}\}_{i=1}^{N}$, \ie, $\{\hat{\mathbf{s}}_i^0\}_{i=1}^{H_0}$, and the mask $\mathbf{m}_i^k$ as input, using a prompt
$$
\mathbf{
a \  photo \  of \  [ENT]
}
$$
to guide the object generation, where the $\mathbf{[ENT]}$ token is expected to be the pseudo label $b_i^k$.

To predict the relation predicates, for each mask pair $(\mathbf{m}_i^k,\mathbf{m}_j^k)$, the ~\moduleDD~ takes $\{\hat{\mathbf{p}}\}_{i=1}^{N}$, the image masks $\mathbf{m}_i^k$ and $\mathbf{m}_j^k$, and the learnable positional embeddings $\mathbf{f}_{sub}$, $\mathbf{f}_{obj}$, $\mathbf{f}_{region}$ as input. For each mask pair, an additional region mask $\mathbf{m}_{r}^k$, \ie,
$$
\mathbf{m}_{r}^k
=
Rec(\mathbf{m}_i^k \cup \mathbf{m}_j^k) - (\mathbf{m}_i^k \cup \mathbf{m}_j^k),
$$
is used to indicate the complement region of the relation, where $Rec$ denotes the enclosing rectangle.
 The $\mathbf{f}_{sub}$, $\mathbf{f}_{obj}$, $\mathbf{f}_{region}$ are added to $\{\hat{\mathbf{p}}\}_{i=1}^{N}$ according to $\mathbf{m}_i^k$, $\mathbf{m}_j^k$, $\mathbf{m}_r^k$ respectively before decoding to indicate the different regions in the image tokens. With the enhanced image tokens and the union mask $\mathbf{m}_i^k \cup \mathbf{m}_j^k \cup \mathbf{m}_{r}^k$, the~\moduleDD~ uses a prompt
$$
\mathbf{
a \  photo \  of \  [SUB]  \ and  \ [OBJ] 
}
$$
$$
\mathbf{
what \  is \  their \  relation \  [REL]
}
$$
to guide the relation generation, where the $\mathbf{[SUB]}$ and $\mathbf{[OBJ]}$ tokens are embedded by the pseudo labels $b_i^k$ and $b_j^k$, and the $\mathbf{[REL]}$ token is expected to be the relation predicate between $(b_i^k,b_j^k)$ with $b_i^k$ as subject and $b_j^k$ as object in the text graph.

Note that to reduce the noise in the pseudo object and relation labels from the caption-parsed text graphs, we change all pseudo labels into their lemma form for the generation.

\subsection{More Details of Inference}

Different from the training procedure, during inference, the framework only takes a scene image $I$ as input without its caption, so that the~\moduleBB~ is not used. With the given target concept sets of object semantics $\mathcal{C}_o$ and relation predicates $\mathcal{C}_r$, the goal for inference is to generate a PSG with its object and relation labels selected from $\mathcal{C}_o$ and $\mathcal{C}_r$.

During inference, an inference stage index $l_{inf}$ is specified to generate the candidate image segments. The model firstly uses the~\moduleAA~ to partition $I$ into $H_{l_{inf}}$ segments $\{\mathbf{s}_i^{l_{inf}}\}_{i=1}^{H_{l_{inf}}}$, which are then merged by the segment merger based on the similarity matrix $\mathbf{Sim}_{l_{inf}}$. Ideally, after swapping rows and columns, $\mathbf{Sim}_{l_{inf}}$ should be a block diagonal matrix in $\{0,1\}^{H_{l_{inf}} \times H_{l_{inf}}}$ with a low rank, and the merging of segments can thus be formulated as a spectral clustering problem. However, $\mathbf{Sim}_{l_{inf}}$ is actually a noisy matrix in $[0,1]^{H_{l_{inf}} \times H_{l_{inf}}}$. To reduce the noise and perform a more accurate clustering, we employ a matrix recovery method~\cite{liu_2012_robust} to recover the low-rank subspace structure of $\mathbf{Sim}_{l_{inf}}$, \ie, by solving a convex optimization problem
$$
\min \limits_{\mathbf{Z}_{l_{inf}},\mathbf{E}_{l_{inf}}} {\|\mathbf{Z}_{l_{inf}}\|_{*} + \lambda \|\mathbf{E}_{l_{inf}}\|_{2,1}},
$$
$$
\text{s.t. } \mathbf{Sim}_{l_{inf}} = \mathbf{Sim}_{l_{inf}} \mathbf{Z}_{l_{inf}} + \mathbf{E}_{l_{inf}},
$$
where $\mathbf{Z}_{l_{inf}}$ denotes the recovered low-rank matrix, $\mathbf{E}_{l_{inf}}$ denotes the noise matrix, $\|\cdot\|_{*}$ denotes the nuclear norm, and $\|\cdot\|_{2,1}$ denotes the $l_{2,1}$ norm. $\lambda$ is a hyperparameter that is set to 0.4 in our experiments.

Then the recovered matrix $\mathbf{Z}_{l_{inf}}$ is applied the normalized cut~\cite{shi_2000_ncut} for clustering, where the segments with similar object semantics tend to be merged into the same cluster. After this step, $D$ merged segmentation masks $\{\hat{\mathbf{m}}_i\}_{i=1}^D$ are obtained.

For each merged mask $\hat{\mathbf{m}}_i$, the~\moduleDD~ uses a similar PET to predict the object label in $\mathcal{C}_o$, which are then be used to predict the relation label in $\mathcal{C}_r$. Different from training, here, the object labels and the relation labels are predicted in a cascaded manner. To select the label in $\mathcal{C}_o$ and $\mathcal{C}_r$, each candidate label is embedded into the prompt (at the $\mathbf{[ENT]}$ or the $\mathbf{[REL]}$ token) to compute its generation probability, which is then used in ranking to select the most probable as the final prediction. Here we use a greedy strategy in implementation to reduce the computation cost.
Following the training procedure, all target concepts in $\mathcal{C}_o$ and $\mathcal{C}_r$ are changed into their lemma form for the generation.

%% file: tex/ap_02_experiment.tex
\section{More Details of Experiments}
\label{sec:supp-experiment}

\subsection{More Details of Datasets for \taskname}

In our experiments, we use the Panoptic Scene Graph dataset~\cite{yang_2022_psg} for the evaluation of the problem \taskname. 
Compared with this dataset, the commonly-used dataset Visual Genome (VG)~\cite{Krishna_2016_VG} has three limitations that make it less suitable for our evaluation.
Firstly, VG only uses bboxes for object location in scene graphs with no fine-grained segmentation masks provided. 
Secondly, the scene graphs in VG are not panoptic, in which only a few objects in the scenes are covered. 
Thirdly, the standard concepts~\cite{xu_2017_messagepass} of object semantics and relation predicates in VG (\ie, 150 objects and 50 relations) are not well-defined enough, where some similar and ambiguous concepts exist, such as $\textit{man}, \textit{men}, \textit{woman}, \textit{person}$ for objects and $\textit{wears}, \textit{wearing}$ for relations. 
In contrast, the Panoptic Scene Graph dataset not only provides object location in the form of both bboxes and segmentation masks, but also contains a more clear, more informative, more coherent class system with comprehensive and panoptic annotations, which is more suitable for the evaluation of \taskname.

The original Panoptic Scene Graph dataset contains 133 object semantics and 56 relation predicates. However, in the original 133 object semantics, there are still some ambiguous classes not well-defined, such as $\textit{window-blind}$ and $\textit{window-other}$, $\textit{floor-wood}$ and $\textit{floor-other-merged}$. To reduce the ambiguity during evaluation, we further merge the ambiguous object semantics with their corresponding annotations, \ie, $\textit{window-blind}$, $\textit{window-other}$ into $\textit{window}$; $\textit{floor-wood}$, $\textit{floor-other-merged}$ into $\textit{floor}$; $\textit{wall-brick}$, $\textit{wall-stone}$, $\textit{wall-tile}$, $\textit{wall-wood}$, $\textit{wall-other-merged}$ into $\textit{wall}$. After merging, 127 object semantics and 56 relation predicates are obtained for our evaluation.

Note that the final set of 127 object semantics consists of 80 thing classes, which represent object classes that can be individually recognized and segmented in an image, and 47 stuff classes, which represent object classes that usually have a homogeneous texture or pattern and are difficult to be segmented individually. In the Panoptic Scene Graph dataset, objects belonging to stuff classes are not segmented individually, with each stuff class having only one mask at most. To accommodate this approach, during the evaluation of our method and the baselines on \taskname, the predicted objects with the same stuff class are merged into a single object.

\subsection{More Details of Baselines for \taskname}

Firstly, we design four baselines that strictly follow the constraints of \taskname~ for a fair comparison. In these baselines, objects in scenes are located by bbox proposals generated by selective search~\cite{uijlings_2013_selective}, which requires no location priors or supervision. For each scene image, 50 proposals are generated.

\begin{itemize}
[align=right,itemindent=0em,labelsep=2pt,labelwidth=1em,leftmargin=*,itemsep=0em] 

\item \textbf{Random} predicts all object semantics and relation predicates fully randomly, where the score for each label is randomly selected from $[0,1]$.

\item \textbf{Prior} augments \textbf{Random} by predicting labels based on the statistical priors in the training set. Specifically, during inference, the model collects the distribution of the target concepts $\mathcal{C}_o$ and $\mathcal{C}_r$ in the training set, then follows the distribution frequency to predict the score in $[0,1]$ for each label.

\item \textbf{MIL} performs the alignment between proposals and textual entities, using a multiple instance learning~\cite{maron_1997_mil} strategy to match the proposals and the entities in captions implicitly. The object label prediction is formulated as a classification problem in a large pre-built vocabulary. Specifically, similar to~\cite{zhong_2021_SGGNLS}, the model builds a large object vocabulary with the most frequent 4,000 entities in the captions in the training set, and the training procedure for object prediction is a 4000-class classification problem. During inference, the model employs WordNet~\cite{miller_1992_wordnet} to match the 4000 classes with the target concepts $\mathcal{C}_o$. Once the object labels are predicted, the relation labels in $\mathcal{C}_r$ are predicted with the statistical prior, similar to \textbf{Prior}.

\item \textbf{SGCLIP} employs the pre-trained CLIP~\cite{radford_2021_clip} to predict both object semantic labels and relation predicate labels. Specifically, for objects, the model uses a prompt 
$$
\mathbf{
a \  photo \  of \  a \  [ENT]
}
$$
to obtain the embedding for each object label in $\mathcal{C}_o$, and assigns the label with the highest cosine similarity to the proposal as the prediction. For relations, the model uses a prompt
$$
\mathbf{
a \  photo \  of \  a \  [SUB] \  [REL] \  a \  [OBJ]
}
$$
to 
obtain the embedding for each relation label in $\mathcal{C}_r$ for each object pair, and assigns the label with the highest cosine similarity as the prediction.

\end{itemize}

By gradually removing the constraints of \taskname, we set two additional baselines to further benchmark the performance of our framework, based on the previous work~\cite{zhong_2021_SGGNLS} for weakly-supervised scene graph generation.

\begin{itemize}
[align=right,itemindent=0em,labelsep=2pt,labelwidth=1em,leftmargin=*,itemsep=0em] 

\item \textbf{SGGNLS-o}~\cite{zhong_2021_SGGNLS} is built without the constraint of no location priors. It extracts object proposals with a detector~\cite{ren_2015_fasterrcnn} pre-trained on OpenImage~\cite{Kuznetsova_2018_OpenImage}. Following~\cite{zhong_2021_SGGNLS}, on average, 36 object proposals are extracted for each image.
It formulates the label prediction as a classification problem within a large pre-built vocabulary, where a 4,000-class object semantics vocabulary and a 1,000-class relation predicate vocabulary are built from the most frequent 4,000 entities and 1,000 relations in the captions in the training set. During inference, the model employs WordNet~\cite{miller_1992_wordnet} to match the 4000 object classes with the target concepts $\mathcal{C}_o$ and 1,000 relation classes with $\mathcal{C}_r$.

\item \textbf{SGGNLS-c}~\cite{zhong_2021_SGGNLS} is built without the constraint of no location priors and no pre-defined concept sets, based on \textbf{SGGNLS-o}. It uses the same proposals as \textbf{SGGNLS-o}. In \textbf{SGGNLS-c}, the target concept sets for inference are known during training. It formulates the label prediction as a classification problem within $\mathcal{C}_o$ and $\mathcal{C}_r$, where all entities and relations from captions in the training set are pre-mapped to $\mathcal{C}_o$ and $\mathcal{C}_r$ through an accurate human-refined mapping as pseudo labels during training.

\end{itemize}

\subsection{More Details of Implementation}

In \modelname, the input image resolution for training is 384$\times$384, and the resolution for inference is 512 for the shortest side. The patch size of the~\moduleAA~ is 16. The filtering threshold in the~\moduleBB~ is set to -0.5. We train \modelname on the COCO Caption dataset~\cite{chen_2015_caption} for 100 epochs.
We use a batch size of 1,728, a learning rate of 0.0001, and the AdamW optimizer~\cite{loshchilov_2017_adamw} with weight decay as 0.05.

%% file: tex/ap_03_result_analysis.tex
\section{More Results on \taskname}
\label{sec:supp-result}

\subsection{More Ablation Studies}

Here we conduct additional ablation studies to further evaluate the effectiveness of two design choices in our framework. 

\noindent\textbf{Positional Embeddings in PET.}
In~\cref{tab:abla-pos}, we compare the different strategies for indicating the different regions in the image tokens in PET. Based on the full PET in \modelname (row 3), we first remove the region embedding $\mathbf{f}_{region}$ (row 2) and further remove the subject embedding $\mathbf{f}_{sub}$ as well as the object embedding $\mathbf{f}_{obj}$ (row 1). The results show that the design of $\mathbf{f}_{sub}$ and $\mathbf{f}_{obj}$ is very important to the generation, without which the model will suffer a significant performance drop. And the design of $\mathbf{f}_{region}$ can further improve the performance by indicating the compliment region information in the image tokens.

\input{table/06_ablation_pos.tex}

\noindent\textbf{Filtering Threshold.}
In~\cref{tab:abla-thresh}, we investigate the effectiveness of setting a filtering threshold $\theta$ to filter out the mismatched image region and caption entity pairs. The results show that compared with the region-entity alignment without filtering (row 1), the introduced $\theta$ (row 2) is simple yet effective in improving the performance significantly.

\input{table/07_ablation_threshold.tex}

\subsection{More Model Diagnosis.}

Here we provide more diagnoses of our framework for a clearer understanding of the efficacy. We answer the following questions.
\textbf{Q1}: How significantly does the pre-trained GroupViT~\cite{xu_2022_group} enhance the learning our framework?
\textbf{Q2}: How does our framework perform with partial ground truth given? 
\textbf{Q3}: How does our framework perform with BLIP~\cite{li_2022_blip} replaced by CLIP~\cite{radford_2021_clip} for the label prediction?

\input{table/08_model_pretrain.tex}

In~\cref{tab:model-pretrain}, we examine the efficacy of the pre-trained GroupViT~\cite{xu_2022_group} in two more training settings: no pre-trained GroupViT weights are used (row 1); initializing weights of GroupViT pre-trained solely on the COCO Caption dataset~\cite{chen_2015_caption} (row 2). The results show that a pre-trained GroupViT is necessary for the effectiveness of our model. Furthermore, GroupViT pre-trained on a large dataset (row 3) can provide very strong location priors and thus facilitates our model significantly (answering \textbf{Q1}).

\input{table/09_partial_gt_clip.tex}
We evaluate the performance of our model on two additional settings with partial ground truth: (i) \textbf{SGCls}, where ground truth object masks are known; (ii) \textbf{PredCls}, where ground truth object masks and semantics are known. The correctness definition is the same as \textbf{SGDet}. The results are shown in~\cref{tab:partial-gt-clip} row 2. The results show that both the segmentation and the relation/entity label prediction still have a large space to improve, especially the label prediction. A better method for label prediction in our challenging setting may improve the performance significantly (answering \textbf{Q2}).

Substituting BLIP with CLIP in our framework for the label prediction, akin to \textbf{PSGCLIP}, results in performance decline across all settings as per~\cref{tab:partial-gt-clip}. The significant drop in \textbf{PredCls} demonstrates CLIP's insensitivity to nuanced relation predicates (answering \textbf{Q3}).

\input{figure/04_more_visualization.tex}

\subsection{More Visualization for Qualitative Evaluation}

We provide more visualization of the predicted PSGs by \modelname in~\cref{fig:more-visualization} for further qualitative evaluation, comparing with the baseline \textbf{SGGNLS-o}.

\subsection{Example of Failure Cases}

Compared with the baseline \textbf{SGGNLS-o}, \cref{fig:more-visualization} shows that our framework is capable of providing more fine-grained labels to each pixel in the image, and is able to reach a panoptic understanding of the scene. However, there are some limitations to our framework that result in some failure cases.

Firstly, the strategy we use to convert the semantic segmentation into instance segmentation is not entirely effective. As shown in~\cref{fig:more-visualization}, our strategy can successfully separate the two cows in (ii), but mistakenly divides the car behind the tree into three parts in (i).

Secondly, our framework faces difficulty in locating small objects in the scene due to limitations in resolution and the grouping strategy for location. As shown in~\cref{fig:more-visualization} (ii) and (iv), our method can identify large objects such as large cows, birds, grass, and sea, but struggles to locate relatively small objects such as small cows in (ii) and people in (iv).

Thirdly, the relation prediction of our framework requires enhancement, as it is not adequately conditioned on the image. As shown in~\cref{fig:more-visualization} (i), the relations between the blue mask of the car and the green mask of the car are predicted as both being \textit{in front of}, which is not reasonable. In this case, \textit{beside} may be a more appropriate prediction (in this case, the first limitation about the segmentation conversion also exists).

%% file: table/06_ablation_pos.tex
\begin{table}[h]
\centering
\resizebox{\linewidth}{!}{
\setlength{\tabcolsep}{8pt}
\begin{tabular}{ccc|cc|cc}
\shline
\multirow{2}{*}{$\mathbf{f}_{sub}$} & \multirow{2}{*}{$\mathbf{f}_{obj}$} & \multirow{2}{*}{$\mathbf{f}_{region}$} & \multicolumn{2}{c|}{PhrDet} & \multicolumn{2}{c}{SGDet} \\
 &  &  & N3R100 & N5R100 & N3R100 & N5R100 \\ 
 \hline
 \ding{56} &  \ding{56} &  \ding{56} & 2.33 & 2.58 & 0.45 & 0.6 \\
 \ding{52} & \ding{52} &  \ding{56} & 10.67 & 11.3 & 2.81 & 3.21 \\
\ding{52} & \ding{52} & \ding{52} & \textbf{12.74} & \textbf{14.37} & \textbf{4.77} & \textbf{5.48} \\
\shline
\end{tabular}}
\caption{\textbf{Ablation Study on Positional Embeddings in PET.} `$\mathbf{f}_{sub}$', `$\mathbf{f}_{obj}$', and `$\mathbf{f}_{region}$' denotes the learnable positional embeddings for indicating the subject region, the object region, and the complement region in the image tokens.}
\label{tab:abla-pos}
\end{table}

%% file: table/07_ablation_threshold.tex
\begin{table}[ht]
\centering
\resizebox{0.8\linewidth}{!}{
\setlength{\tabcolsep}{8pt}
\begin{tabular}{c|cc|cc}
\shline
\multirow{2}{*}{Thresh} & \multicolumn{2}{c|}{PhrDet} & \multicolumn{2}{c}{SGDet} \\
  & N3R100 & N5R100 & N3R100 & N5R100 \\ 
 \hline
\ding{56} & 10.39 & 10.8 & 3.09 & 3.19 \\
\ding{52} & \textbf{12.74} & \textbf{14.37} & \textbf{4.77} & \textbf{5.48} \\
\shline
\end{tabular}}
\caption{\textbf{Ablation Study on Filtering Threshold.} `Thresh' denotes the filtering threshold $\theta$ for filtering out the mismatched image region and caption entity pairs.}
\label{tab:abla-thresh}
\end{table}

%% file: table/08_model_pretrain.tex
\begin{table}[h]
    \centering
    \resizebox{\linewidth}{!}{
    \setlength{\tabcolsep}{8pt}
    \begin{tabular}{c|cc|cc}
        \shline
        Pre-trained & \multicolumn{2}{c|}{PhrDet} & \multicolumn{2}{c}{SGDet} \\
        Weights & N3R100 & N5R100 & N3R100 & N5R100 \\ 
         \hline
        \ding{56} & 0 & 0 & 0 & 0 \\
        COCO Caption~\cite{chen_2015_caption} & 1.99 & 2.51 & 0.07 & 0.1 \\
        CC12M~\cite{changpinyo_2021_cc12m, sharma-etal-2018-conceptual}+YFCC~\cite{thomee_2016_yfcc100m} & \textbf{12.74} & \textbf{14.37} & \textbf{4.77} & \textbf{5.48} \\
        \shline
    \end{tabular}}
    \caption{\textbf{Examination on Pre-trained GroupViT Weights.}}
    \label{tab:model-pretrain}
\end{table}

%% file: table/09_partial_gt_clip.tex
\begin{table}[h]
    \centering
    \resizebox{1\linewidth}{!}{
    \setlength{\tabcolsep}{3pt}
    \begin{tabular}{c|cc|cc|cc}
        \shline
        \multirow{2}{*}{Method} & \multicolumn{2}{c|}{SGCls} & \multicolumn{2}{c|}{PredCls} & \multicolumn{2}{c}{SGDet} \\
         & N3R100 & N5R100 & N3R100 & N5R100 & N3R100 & N5R100 \\ 
         \hline
        PSGCLIP & 7.38 & 9.11 & 25.72 & 26.16 & 2.83 & 3.23 \\
        Ours & \textbf{9.51} & \textbf{10.79} & \textbf{36.28} & \textbf{39.79} & \textbf{4.77} & \textbf{5.48} \\
        \shline
    \end{tabular}}
    \caption{\textbf{More Evaluation Settings.}}
    \label{tab:partial-gt-clip}
\end{table}

%% file: figure/04_more_visualization.tex
\begin{figure*}[t]
  \centering
  \includegraphics[width=0.985\linewidth]{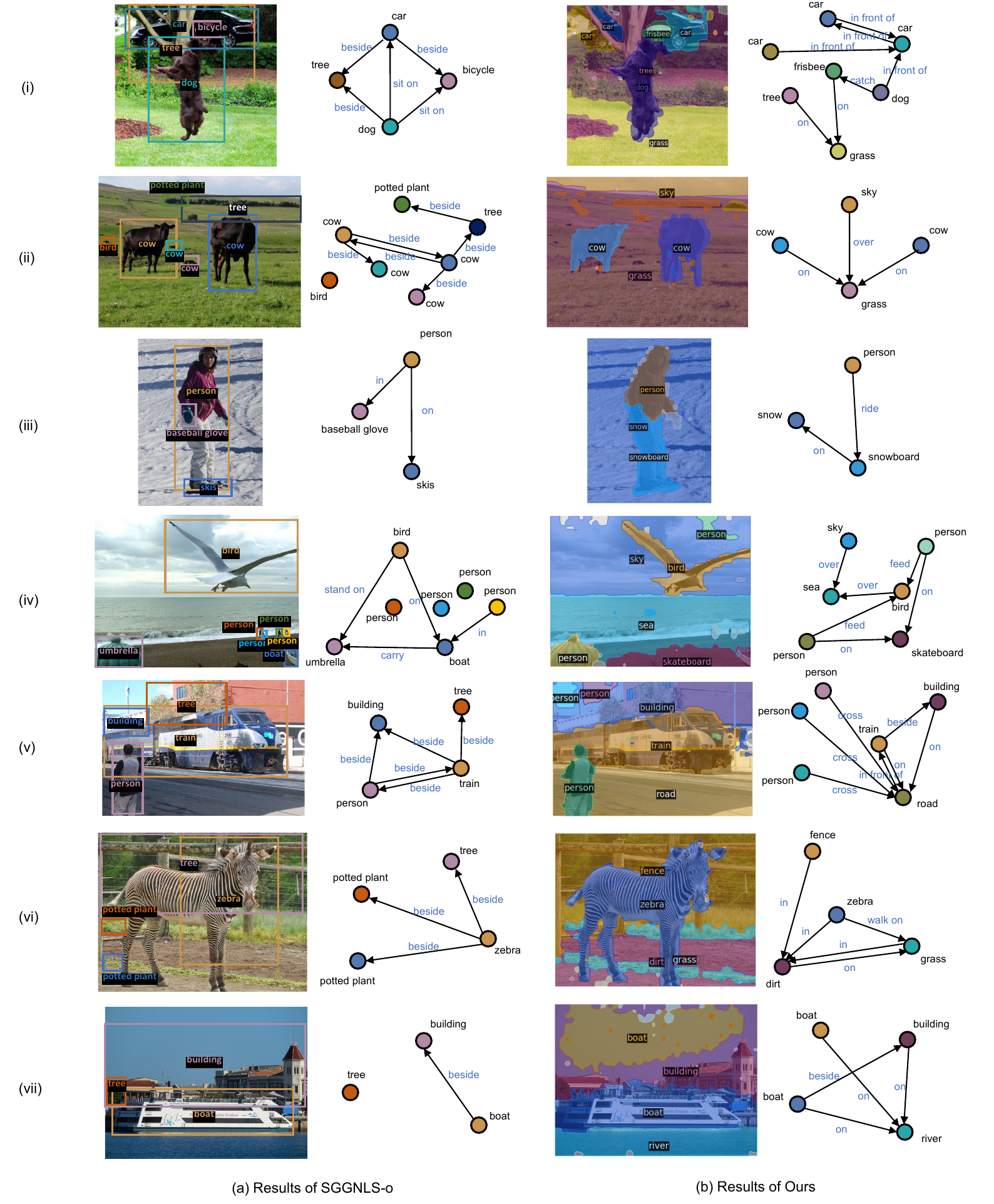}
   \caption{\textbf{More Qualitative Comparison between SGGNLS-o (a) and Ours (b).} For each method, the results of object location are shown on the left, while the results of scene graph generation are shown on the right. For \textbf{SGGNLS-o} and \textbf{Ours}, the visualized relations are picked from the top 10 triplets in the scene graph (the predicate score should be greater than 0.6). For \textbf{SGGNLS-o}, only proposals matched with ground truth (only requires a correct location, ignores the semantics) are visualized. 
   }
   \label{fig:more-visualization}
\end{figure*}

%% file: main.bbl
\begin{thebibliography}{10}\itemsep=-1pt

\bibitem{aditya_2017_graph}
Somak Aditya, Yezhou Yang, Chitta Baral, Yiannis Aloimonos, and Cornelia
  Fermüller.
\newblock Image understanding using vision and reasoning through scene
  description graph.
\newblock {\em Computer Vision and Image Understanding}, 2017.

\bibitem{Amiri_2022_ReasoningWS}
S. Amiri, Kishan Chandan, and Shiqi Zhang.
\newblock Reasoning with scene graphs for robot planning under partial
  observability.
\newblock {\em IEEE Robotics and Automation Letters}, 7:5560--5567, 2022.

\bibitem{Angeli_2015_OpenIE}
Gabor Angeli, Melvin Johnson, and Christopher~D. Manning.
\newblock Leveraging linguistic structure for open domain information
  extraction.
\newblock In {\em Annual Meeting of the Association for Computational
  Linguistics}, 2015.

\bibitem{changpinyo_2021_cc12m}
Soravit Changpinyo, Piyush Sharma, Nan Ding, and Radu Soricut.
\newblock {Conceptual 12M}: Pushing web-scale image-text pre-training to
  recognize long-tail visual concepts.
\newblock In {\em CVPR}, 2021.

\bibitem{chen_2015_caption}
Xinlei Chen, Hao Fang, Tsung-Yi Lin, Ramakrishna Vedantam, Saurabh Gupta, Piotr
  Dollar, and C.~Lawrence Zitnick.
\newblock Microsoft {COCO} captions: Data collection and evaluation server.
\newblock {\em arXiv preprint arXiv:1504.00325}, 2015.

\bibitem{chen19acl}
Zhenfang Chen, Lin Ma, Wenhan Luo, and Kwan-Yee~K Wong.
\newblock Weakly-supervised spatio-temporally grounding natural sentence in
  video.
\newblock In {\em Proc. 57th Annual Meeting of the Association for
  Computational Linguistics}, 2019.

\bibitem{zfchen2021iclr}
Zhenfang Chen, Jiayuan Mao, Jiajun Wu, Kwan-Yee~K Wong, Joshua~B. Tenenbaum,
  and Chuang Gan.
\newblock Grounding physical concepts of objects and events through dynamic
  visual reasoning.
\newblock In {\em International Conference on Learning Representations}, 2021.

\bibitem{chen2020cops}
Zhenfang Chen, Peng Wang, Lin Ma, Kwan-Yee~K Wong, and Qi Wu.
\newblock Cops-ref: A new dataset and task on compositional referring
  expression comprehension.
\newblock In {\em Proceedings of the IEEE/CVF Conference on Computer Vision and
  Pattern Recognition}, pages 10086--10095, 2020.

\bibitem{dhamo_2020_simsg}
Helisa Dhamo, Azade Farshad, Iro Laina, Nassir Navab, Gregory~D. Hager,
  Federico Tombari, and Christian Rupprecht.
\newblock Semantic image manipulation using scene graphs.
\newblock In {\em CVPR}, 2020.

\bibitem{dong2022maskclip}
Xiaoyi Dong, Yinglin Zheng, Jianmin Bao, Ting Zhang, Dongdong Chen, Hao Yang,
  Ming Zeng, Weiming Zhang, Lu Yuan, Dong Chen, et~al.
\newblock Maskclip: Masked self-distillation advances contrastive
  language-image pretraining.
\newblock {\em arXiv preprint arXiv:2208.12262}, 2022.

\bibitem{Gadre_2022_ContinuousSR}
Samir~Yitzhak Gadre, Kiana Ehsani, Shuran Song, and Roozbeh Mottaghi.
\newblock Continuous scene representations for embodied ai.
\newblock {\em 2022 IEEE/CVF Conference on Computer Vision and Pattern
  Recognition (CVPR)}, pages 14829--14839, 2022.

\bibitem{gan2017vqs}
Chuang Gan, Yandong Li, Haoxiang Li, Chen Sun, and Boqing Gong.
\newblock Vqs: Linking segmentations to questions and answers for supervised
  attention in vqa and question-focused semantic segmentation.
\newblock In {\em Proceedings of the IEEE international conference on computer
  vision}, pages 1811--1820, 2017.

\bibitem{Gu_2019_external}
Jiuxiang Gu, Handong Zhao, Zhe Lin, Sheng Li, Jianfei Cai, and Mingyang Ling.
\newblock Scene graph generation with external knowledge and image
  reconstruction.
\newblock In {\em Proceedings of the IEEE/CVF Conference on Computer Vision and
  Pattern Recognition (CVPR)}, June 2019.

\bibitem{hudson_2019_gqa}
Drew~A Hudson and Christopher~D Manning.
\newblock Gqa: A new dataset for real-world visual reasoning and compositional
  question answering.
\newblock In {\em Proceedings of the IEEE/CVF conference on computer vision and
  pattern recognition}, pages 6700--6709, 2019.

\bibitem{johnson_2018_image}
Justin Johnson, Agrim Gupta, and Li Fei-Fei.
\newblock Image generation from scene graphs.
\newblock In {\em CVPR}, 2018.

\bibitem{karpathy2015deep}
Andrej Karpathy and Li Fei-Fei.
\newblock Deep visual-semantic alignments for generating image descriptions.
\newblock In {\em Proceedings of the IEEE conference on computer vision and
  pattern recognition}, pages 3128--3137, 2015.

\bibitem{kazemzadeh2014referitgame}
Sahar Kazemzadeh, Vicente Ordonez, Mark Matten, and Tamara Berg.
\newblock Referitgame: Referring to objects in photographs of natural scenes.
\newblock In {\em Proceedings of the 2014 conference on empirical methods in
  natural language processing (EMNLP)}, pages 787--798, 2014.

\bibitem{Khandelwal_2021_SegmSG}
Siddhesh Khandelwal, Mohammed Suhail, and Leonid Sigal.
\newblock Segmentation-grounded scene graph generation.
\newblock {\em 2021 IEEE/CVF International Conference on Computer Vision
  (ICCV)}, pages 15859--15869, 2021.

\bibitem{Krishna_2016_VG}
Ranjay Krishna, Yuke Zhu, Oliver Groth, Justin Johnson, Kenji Hata, Joshua
  Kravitz, Stephanie Chen, Yannis Kalantidis, Li-Jia Li, David~A. Shamma,
  Michael~S. Bernstein, and Fei-Fei Li.
\newblock {Visual Genome}: Connecting language and vision using crowdsourced
  dense image annotations.
\newblock {\em International Journal of Computer Vision}, 123:32--73, 2016.

\bibitem{Kuznetsova_2018_OpenImage}
Alina Kuznetsova, Hassan Rom, Neil~Gordon Alldrin, Jasper R.~R. Uijlings, Ivan
  Krasin, Jordi Pont-Tuset, Shahab Kamali, Stefan Popov, Matteo Malloci,
  Alexander Kolesnikov, Tom Duerig, and Vittorio Ferrari.
\newblock The open images dataset v4.
\newblock {\em International Journal of Computer Vision}, 128:1956--1981, 2018.

\bibitem{li2022language}
Boyi Li, Kilian~Q Weinberger, Serge Belongie, Vladlen Koltun, and Ren{\'e}
  Ranftl.
\newblock Language-driven semantic segmentation.
\newblock {\em arXiv preprint arXiv:2201.03546}, 2022.

\bibitem{li_2022_blip}
Junnan Li, Dongxu Li, Caiming Xiong, and Steven Hoi.
\newblock Blip: Bootstrapping language-image pre-training for unified
  vision-language understanding and generation.
\newblock In {\em ICML}, 2022.

\bibitem{Li_2022_IntegratingOA}
Xingchen Li, Long Chen, Wenbo Ma, Yi Yang, and Jun Xiao.
\newblock Integrating object-aware and interaction-aware knowledge for weakly
  supervised scene graph generation.
\newblock {\em Proceedings of the 30th ACM International Conference on
  Multimedia}, 2022.

\bibitem{Lin_2014_COCO}
Tsung-Yi Lin, Michael Maire, Serge~J. Belongie, James Hays, Pietro Perona, Deva
  Ramanan, Piotr Doll{\'a}r, and C.~Lawrence Zitnick.
\newblock Microsoft coco: Common objects in context.
\newblock In {\em European Conference on Computer Vision}, 2014.

\bibitem{Lin_2020_gpsnet}
Xin Lin, Changxing Ding, Jinquan Zeng, and Dacheng Tao.
\newblock Gps-net: Graph property sensing network for scene graph generation.
\newblock In {\em Proceedings of the IEEE/CVF Conference on Computer Vision and
  Pattern Recognition (CVPR)}, June 2020.

\bibitem{liu_2012_robust}
Guangcan Liu, Zhouchen Lin, Shuicheng Yan, Ju Sun, Yong Yu, and Yi Ma.
\newblock Robust recovery of subspace structures by low-rank representation.
\newblock {\em IEEE transactions on pattern analysis and machine intelligence},
  35(1):171--184, 2012.

\bibitem{loshchilov_2017_adamw}
Ilya Loshchilov and Frank Hutter.
\newblock Decoupled weight decay regularization.
\newblock {\em arXiv preprint arXiv:1711.05101}, 2017.

\bibitem{luddecke2022image}
Timo L{\"u}ddecke and Alexander Ecker.
\newblock Image segmentation using text and image prompts.
\newblock In {\em Proceedings of the IEEE/CVF Conference on Computer Vision and
  Pattern Recognition}, pages 7086--7096, 2022.

\bibitem{luo2022segclip}
Huaishao Luo, Junwei Bao, Youzheng Wu, Xiaodong He, and Tianrui Li.
\newblock Segclip: Patch aggregation with learnable centers for open-vocabulary
  semantic segmentation.
\newblock {\em arXiv preprint arXiv:2211.14813}, 2022.

\bibitem{manning_2014_corenlp}
Christopher~D. Manning, Mihai Surdeanu, John Bauer, Jenny Finkel, Steven~J.
  Bethard, and David McClosky.
\newblock The {Stanford} {CoreNLP} natural language processing toolkit.
\newblock In {\em Association for Computational Linguistics (ACL) System
  Demonstrations}, pages 55--60, 2014.

\bibitem{mao2016generation}
Junhua Mao, Jonathan Huang, Alexander Toshev, Oana Camburu, Alan~L Yuille, and
  Kevin Murphy.
\newblock Generation and comprehension of unambiguous object descriptions.
\newblock In {\em Proceedings of the IEEE conference on computer vision and
  pattern recognition}, pages 11--20, 2016.

\bibitem{maron_1997_mil}
Oded Maron and Tom{\'a}s Lozano-P{\'e}rez.
\newblock A framework for multiple-instance learning.
\newblock {\em Advances in neural information processing systems}, 10, 1997.

\bibitem{miller_1992_wordnet}
George~A. Miller.
\newblock {W}ord{N}et: A lexical database for {E}nglish.
\newblock In {\em Speech and Natural Language: Proceedings of a Workshop Held
  at Harriman, New York, {F}ebruary 23-26, 1992}, 1992.

\bibitem{Peyre_2017_WSVR}
Julia Peyre, Ivan Laptev, Cordelia Schmid, and Josef Sivic.
\newblock Weakly-supervised learning of visual relations.
\newblock {\em 2017 IEEE International Conference on Computer Vision (ICCV)},
  pages 5189--5198, 2017.

\bibitem{plummer2015flickr30k}
Bryan~A Plummer, Liwei Wang, Chris~M Cervantes, Juan~C Caicedo, Julia
  Hockenmaier, and Svetlana Lazebnik.
\newblock Flickr30k entities: Collecting region-to-phrase correspondences for
  richer image-to-sentence models.
\newblock In {\em Proceedings of the IEEE international conference on computer
  vision}, pages 2641--2649, 2015.

\bibitem{radford_2021_clip}
Alec Radford, Jong~Wook Kim, Chris Hallacy, Aditya Ramesh, Gabriel Goh,
  Sandhini Agarwal, Girish Sastry, Amanda Askell, Pamela Mishkin, Jack Clark,
  et~al.
\newblock Learning transferable visual models from natural language
  supervision.
\newblock In {\em International conference on machine learning}, pages
  8748--8763, 2021.

\bibitem{ren_2015_fasterrcnn}
Shaoqing Ren, Kaiming He, Ross Girshick, and Jian Sun.
\newblock Faster r-cnn: Towards real-time object detection with region proposal
  networks.
\newblock {\em Advances in neural information processing systems}, 28, 2015.

\bibitem{rohrbach2016grounding}
Anna Rohrbach, Marcus Rohrbach, Ronghang Hu, Trevor Darrell, and Bernt Schiele.
\newblock Grounding of textual phrases in images by reconstruction.
\newblock In {\em Computer Vision--ECCV 2016: 14th European Conference,
  Amsterdam, The Netherlands, October 11--14, 2016, Proceedings, Part I 14},
  pages 817--834. Springer, 2016.

\bibitem{Schuster_2015_GeneratingSP}
Sebastian Schuster, Ranjay Krishna, Angel~X. Chang, Li Fei-Fei, and
  Christopher~D. Manning.
\newblock Generating semantically precise scene graphs from textual
  descriptions for improved image retrieval.
\newblock In {\em VL@EMNLP}, 2015.

\bibitem{sharma-etal-2018-conceptual}
Piyush Sharma, Nan Ding, Sebastian Goodman, and Radu Soricut.
\newblock Conceptual captions: A cleaned, hypernymed, image alt-text dataset
  for automatic image captioning.
\newblock In {\em Proceedings of the 56th Annual Meeting of the Association for
  Computational Linguistics (Volume 1: Long Papers)}, pages 2556--2565, July
  2018.

\bibitem{shi_2000_ncut}
Jianbo Shi and J. Malik.
\newblock Normalized cuts and image segmentation.
\newblock {\em IEEE Transactions on Pattern Analysis and Machine Intelligence},
  22(8):888--905, 2000.

\bibitem{shi_2019_explainable}
Jiaxin Shi, Hanwang Zhang, and Juanzi Li.
\newblock Explainable and explicit visual reasoning over scene graphs.
\newblock In {\em Proceedings of the IEEE/CVF conference on computer vision and
  pattern recognition}, pages 8376--8384, 2019.

\bibitem{Shi_2021_ASB}
Jing Shi, Yiwu Zhong, Ning Xu, Yin Li, and Chenliang Xu.
\newblock A simple baseline for weakly-supervised scene graph generation.
\newblock {\em 2021 IEEE/CVF International Conference on Computer Vision
  (ICCV)}, pages 16373--16382, 2021.

\bibitem{Tang_2020_unbias}
Kaihua Tang, Yulei Niu, Jianqiang Huang, Jiaxin Shi, and Hanwang Zhang.
\newblock Unbiased scene graph generation from biased training.
\newblock In {\em Proceedings of the IEEE/CVF Conference on Computer Vision and
  Pattern Recognition (CVPR)}, June 2020.

\bibitem{teney_2017_graph}
Damien Teney, Lingqiao Liu, and Anton van~den Hengel.
\newblock Graph-structured representations for visual question answering.
\newblock In {\em Proceedings of the IEEE Conference on Computer Vision and
  Pattern Recognition (CVPR)}, July 2017.

\bibitem{thomee_2016_yfcc100m}
Bart Thomee, David~A Shamma, Gerald Friedland, Benjamin Elizalde, Karl Ni,
  Douglas Poland, Damian Borth, and Li-Jia Li.
\newblock Yfcc100m: The new data in multimedia research.
\newblock {\em Communications of the ACM}, 59(2):64--73, 2016.

\bibitem{uijlings_2013_selective}
Jasper~RR Uijlings, Koen~EA Van De~Sande, Theo Gevers, and Arnold~WM Smeulders.
\newblock Selective search for object recognition.
\newblock {\em International journal of computer vision}, 104:154--171, 2013.

\bibitem{Vaswani_2017_AttentionIA}
Ashish Vaswani, Noam~M. Shazeer, Niki Parmar, Jakob Uszkoreit, Llion Jones,
  Aidan~N. Gomez, Lukasz Kaiser, and Illia Polosukhin.
\newblock Attention is all you need.
\newblock In {\em NIPS}, 2017.

\bibitem{Wang_2023_PSGWorkshop}
Qixun Wang, Xiaofeng Guo, and Haofan Wang.
\newblock 1st place solution for psg competition with eccv'22 sensehuman
  workshop.
\newblock {\em ArXiv}, abs/2302.02651, 2023.

\bibitem{Wu_2019_UnifiedVE}
Hao Wu, Jiayuan Mao, Yufeng Zhang, Yuning Jiang, Lei Li, Weiwei Sun, and
  Wei-Ying Ma.
\newblock Unified visual-semantic embeddings: Bridging vision and language with
  structured meaning representations.
\newblock {\em 2019 IEEE/CVF Conference on Computer Vision and Pattern
  Recognition (CVPR)}, pages 6602--6611, 2019.

\bibitem{xu_2017_messagepass}
Danfei Xu, Yuke Zhu, Christopher~B Choy, and Li Fei-Fei.
\newblock Scene graph generation by iterative message passing.
\newblock In {\em Proceedings of the IEEE conference on computer vision and
  pattern recognition}, pages 5410--5419, 2017.

\bibitem{xu_2022_group}
Jiarui Xu, Shalini De~Mello, Sifei Liu, Wonmin Byeon, Thomas Breuel, Jan Kautz,
  and Xiaolong Wang.
\newblock {GroupViT}: Semantic segmentation emerges from text supervision.
\newblock In {\em Proceedings of the IEEE/CVF Conference on Computer Vision and
  Pattern Recognition (CVPR)}, pages 18134--18144, June 2022.

\bibitem{yang_2022_psg}
Jingkang Yang, Yi~Zhe Ang, Zujin Guo, Kaiyang Zhou, Wayne Zhang, and Ziwei Liu.
\newblock Panoptic scene graph generation.
\newblock In {\em ECCV}, 2022.

\bibitem{Yang_2018_GraphRCNN}
Jianwei Yang, Jiasen Lu, Stefan Lee, Dhruv Batra, and Devi Parikh.
\newblock Graph r-cnn for scene graph generation.
\newblock In {\em Proceedings of the European Conference on Computer Vision
  (ECCV)}, September 2018.

\bibitem{yao_2022_filip}
Lewei Yao, Runhui Huang, Lu Hou, Guansong Lu, Minzhe Niu, Hang Xu, Xiaodan
  Liang, Zhenguo Li, Xin Jiang, and Chunjing Xu.
\newblock {FILIP}: Fine-grained interactive language-image pre-training.
\newblock In {\em International Conference on Learning Representations}, 2022.

\bibitem{Ye_2021_Linguistic}
Keren Ye and Adriana Kovashka.
\newblock Linguistic structures as weak supervision for visual scene graph
  generation.
\newblock In {\em Proceedings of the IEEE/CVF Conference on Computer Vision and
  Pattern Recognition (CVPR)}, June 2021.

\bibitem{yu2018mattnet}
Licheng Yu, Zhe Lin, Xiaohui Shen, Jimei Yang, Xin Lu, Mohit Bansal, and
  Tamara~L Berg.
\newblock Mattnet: Modular attention network for referring expression
  comprehension.
\newblock In {\em Proceedings of the IEEE conference on computer vision and
  pattern recognition}, pages 1307--1315, 2018.

\bibitem{yu2016modeling}
Licheng Yu, Patrick Poirson, Shan Yang, Alexander~C Berg, and Tamara~L Berg.
\newblock Modeling context in referring expressions.
\newblock In {\em Computer Vision--ECCV 2016: 14th European Conference,
  Amsterdam, The Netherlands, October 11-14, 2016, Proceedings, Part II 14},
  pages 69--85. Springer, 2016.

\bibitem{Zareian_2020_WSVSP}
Alireza Zareian, Svebor Karaman, and Shih-Fu Chang.
\newblock Weakly supervised visual semantic parsing.
\newblock {\em 2020 IEEE/CVF Conference on Computer Vision and Pattern
  Recognition (CVPR)}, pages 3733--3742, 2020.

\bibitem{zeng2020dense}
Runhao Zeng, Haoming Xu, Wenbing Huang, Peihao Chen, Mingkui Tan, and Chuang
  Gan.
\newblock Dense regression network for video grounding.
\newblock In {\em Proceedings of the IEEE/CVF Conference on Computer Vision and
  Pattern Recognition}, pages 10287--10296, 2020.

\bibitem{zhang_2017_ppr}
Hanwang Zhang, Zawlin Kyaw, Jinyang Yu, and Shih-Fu Chang.
\newblock Ppr-fcn: Weakly supervised visual relation detection via parallel
  pairwise r-fcn.
\newblock In {\em Proceedings of the IEEE international conference on computer
  vision}, pages 4233--4241, 2017.

\bibitem{zhong_2021_SGGNLS}
Yiwu Zhong, Jing Shi, Jianwei Yang, Chenliang Xu, and Yin Li.
\newblock Learning to generate scene graph from natural language supervision.
\newblock In {\em ICCV}, 2021.

\bibitem{zhou2022extract}
Chong Zhou, Chen~Change Loy, and Bo Dai.
\newblock Extract free dense labels from clip.
\newblock In {\em Computer Vision--ECCV 2022: 17th European Conference, Tel
  Aviv, Israel, October 23--27, 2022, Proceedings, Part XXVIII}, pages
  696--712. Springer, 2022.

\end{thebibliography}
